\documentclass[lettersize,journal]{IEEEtran}
\usepackage{amsmath,amsfonts}
\usepackage{algorithm}
\usepackage{algorithmicx}
\usepackage{algpseudocode}
\usepackage{array}
\usepackage[caption=false,font=normalsize,labelfont=sf,textfont=sf]{subfig}
 \usepackage{amssymb}
\usepackage{textcomp}
\usepackage{stfloats}
\usepackage{url}
\usepackage{verbatim}
\usepackage{graphicx}
\usepackage{cite}
\usepackage{color}
\usepackage{hyperref}
\usepackage{import}

\usepackage{todonotes}

\usepackage{amsmath}

\newcolumntype{P}[1]{>{\centering\arraybackslash}p{#1}}
\newcolumntype{C}[1]{>{\centering\arraybackslash}m{#1}}

\def \RED [#1]{\color{red}#1 \color{black}}
\def \textHT [#1]{\color{red}$\mathbf{#1}$\color{black}}
\def \textLT [#1]{\color{blue}$\mathbf{#1}$\color{black}}

\begin{document}
\title{AutoMerge: A Framework for Map Assembling \\ and Smoothing in City-scale Environments}

\author{Peng Yin\textsuperscript{1,*},~\IEEEmembership{Member,~IEEE},
        Shiqi Zhao\textsuperscript{2},
        Haowen Lai\textsuperscript{3},
        Ruohai Ge\textsuperscript{1},\\
        Ji Zhang\textsuperscript{1},
        Howie Choset\textsuperscript{1}, ~\IEEEmembership{Fellow,~IEEE},
        and Sebastian Scherer\textsuperscript{1}, ~\IEEEmembership{Senior Member,~IEEE}
\thanks{Peng Yin and Shiqi Zhao are with the Department of Mechanical Engineering, City University of Hong Kong, Hong Kong 518057, China. {\{pengyin@cityu.edu.hk, ryanzhao9459@gmail.com}\}.}
\thanks{Ruohai Ge, Ji Zhang, Howie Choset, and Sebastian Scherer are with the Robotics Institute, Carnegie Mellon University, Pittsburgh, PA 15213, USA. {(ruohaig, zhangji, choset, basti)@andrew.cmu.edu}.}
\thanks{Haowen Lai is with the Department of Automation, Tsinghua University, Beijing, China. {(lhw19@mails.tsinghua.edu.cn)}}
\thanks{\textsuperscript{*}Corresponding author: Peng Yin (pengyin@cityu.edu.hk)}
}

\markboth{IEEE Transactions on Robotics (T-RO). 
Preprint Version. July 2023}
{Yin \MakeLowercase{\textit{et al.}}: AutoMerge: A Framework for Map Assembling and Smoothing in City-scale Environments}


%

\maketitle
\begin{abstract}

In the era of advancing autonomous driving and increasing reliance on geospatial information, high-precision mapping not only demands accuracy but also flexible construction.
Current approaches mainly rely on expensive mapping devices, which are time-consuming for city-scale map construction and vulnerable to erroneous data associations without accurate GPS assistance.
We present AutoMerge, a novel framework for merging large-scale maps that surpasses these limitations, which  (i) provides robust place recognition performance despite differences in both translation and viewpoint, (ii) is capable of identifying and discarding incorrect loop closures caused by perceptual aliasing, and (iii) effectively associates and optimizes large-scale and numerous map segments in the real-world scenario.
AutoMerge utilizes multi-perspective fusion and adaptive loop closure detection for accurate data associations, and it uses incremental merging to assemble large maps from individual trajectory segments given in random order and with no initial estimations. 
Furthermore, AutoMerge performs pose-graph optimization after assembling the segments to smooth the merged map globally.
We demonstrate AutoMerge on both city-scale merging (120km) and campus-scale repeated merging (4.5km$\times$8). 
The experiments show that AutoMerge (i) surpasses the second-and third-best methods by 0.9\% and 6.5\% recall in segment retrieval, (ii) achieves comparable 3D mapping accuracy for 120 km large-scale map assembly, (iii), and it is robust to temporally-spaced revisits. 
To our knowledge, AutoMerge is the first mapping approach to merge hundreds of kilometers of individual segments without using GPS. 
\end{abstract}

\begin{IEEEkeywords}
Map Merging, Viewpoint-invariant Localization, Multi-agent SLAM, Incremental Mapping, GPS-denied
\end{IEEEkeywords}

\begingroup
\let\clearpage\relax

\section{Introduction}
\label{sec:introduction}

    \begin{figure}[t]
        \begin{center}
          \includegraphics[width=\linewidth]{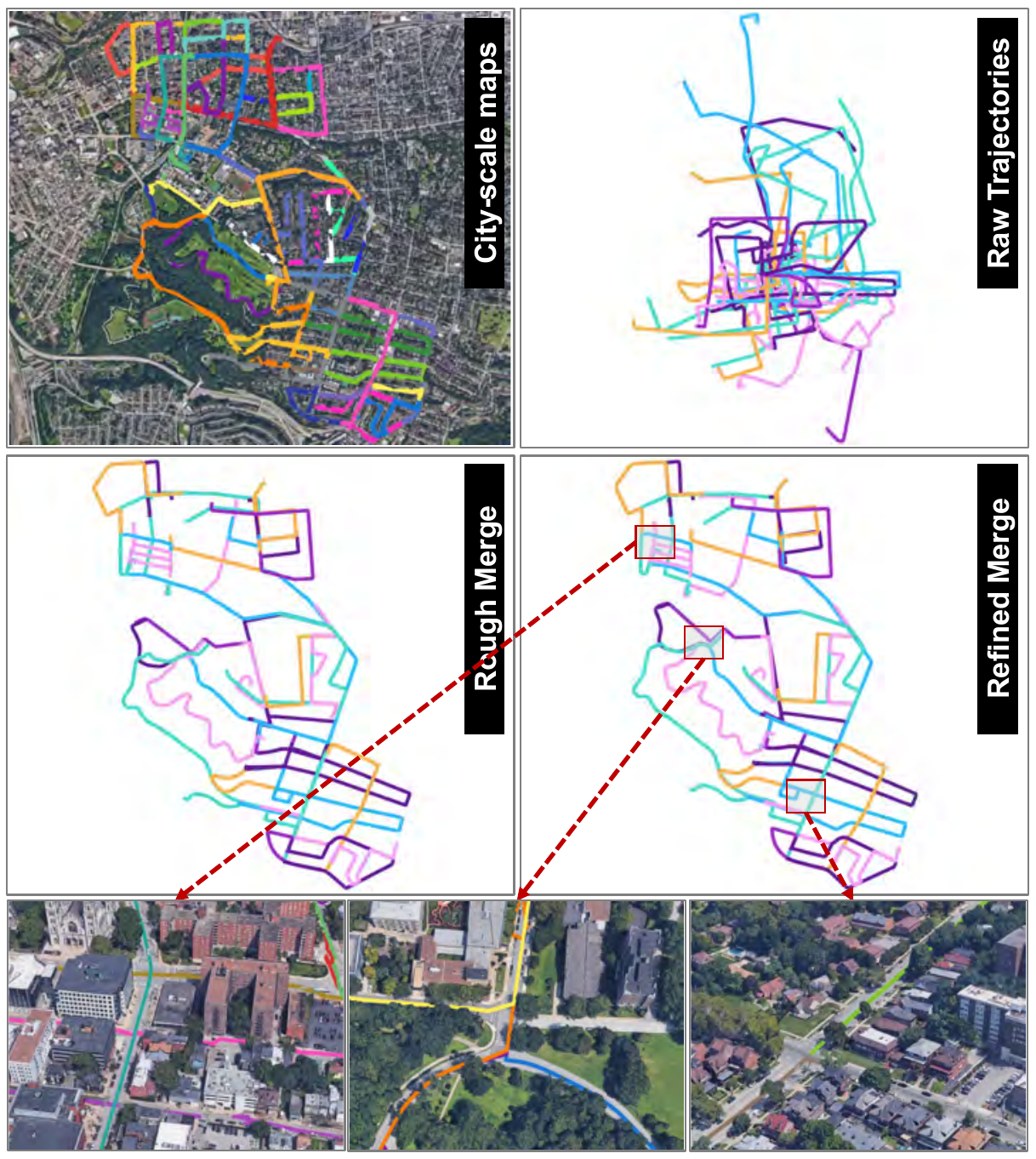}
        \end{center}
        \caption{\textbf{Map Merging in the City of Pittsburgh.}
        Map merging for $50$ segments in the City of Pittsburgh, Pennsylvania, USA.
        The mapped areas include Shadyside, Squirrel Hill, Bloomfield, etc, totaling $180km$ in distance.
        AutoMerge can associate their inner connections with our proposed adaptive loop closure detection with an incorrect matches rejection mechanism, and achieve global mapping through rough/refined merging procedures.
        }
        \label{fig:idea}
    \end{figure}

    \IEEEPARstart{L}{arge}-scale 3D mapping is one of the fundamental topics in robotics research due to its capacity to provide accurate localization and 3D environment representation for high-level perception and planning tasks.
    Moreover, as autonomous driving technology advances and becomes increasingly prevalent, the need for precise and up-to-date crowdsourced maps is crucial for ensuring safe and efficient navigation.
    For both single- and multi-agent mapping systems, merging coupled segments into the same world coordinates becomes necessary to generate accurate localization and mapping results.
    However, as shown in Fig.~\ref{fig:idea}, accurate and robust data associations among a large number of different map segments, especially in large-scale environments, is still a challenging problem.
    Factors that contribute to the difficulty of this problem include:
    \begin{itemize}
        \item Loop closures are susceptible to both translation and orientation differences; when re-visiting the same area, the place descriptor will vary under different perspectives.
        \item Different areas may share similar geometries, such as a long street, a highway, etc., which may cause incorrect data associations among segments without overlaps.
        \item Most methods are extremely sensitive to failed matches; even a few incorrect data associations between segments can turn the global map optimization into an ill-posed problem. 
    \end{itemize}
    
    \noindent Traditional map merging approaches rely on good initialization~\cite{Merge:LAMP} and accurate odometry estimation.
    For a single agent revisiting an area or a multi-agent system without prior knowledge, previous works find matches based on a single frame and utilize RANSAC~\cite{lpr:ransac} to filter out incorrect matches via Euclidean constraints.
    However, such methods rely highly on the discriminability of unique areas, which is hard to guarantee in both scenarios mentioned above. 
    
    In this work, we propose a 3D mapping system, AutoMerge, that enables robust and accurate large-scale map merging under significant viewpoint differences.
    AutoMerge can provide reliable loop closures for an initial rough alignment (i.e., rigid transformation matrix) between different trajectories and then perform global reformation for all the incoming trajectories.
    For either single-agent segment revisiting or multi-agent collaboration mapping, perspective differences and incomplete observations can occur in the 3D representations for the same area, and traditional descriptors~\cite{PR:pointnetvlad, PR:overlapnet} are sensitive to the above scenarios.
    In general, a point-/voxel-based feature extraction approach is designed to be translation-invariant; however, the local viewpoint differences (especially in a detour) can affect the extracted local features.
    In our previous work, we notice that spherical projection shows advanced performance under orientation differences~\cite{spherevlad}, and multi-perspective fusion\cite{fusionvlad} can provide robustness against viewpoint differences.
    In AutoMerge, we develop a novel multi-perspective fusion-based approach for a 3D place descriptor, which combines different perspective-invariant properties \cite{fusionvlad, spherevlad} by leveraging the network features with additional attention-based~\cite{attention_all_your_need} fusion layers.
    As a result of the new place descriptor, AutoMerge can provide the highest average recall rates and lowest false positive rates compared to other state-of-the-art methods.
    Our descriptor can be extracted in real-time, making it suitable for both accurate offline merging for map refinement, and fast incremental merging for multi-agent mapping.

    Perceptual aliasing caused by similar scenarios (long streets, crossroads, highways, etc.) often results in incorrect data association.
    Such failures can cause catastrophic problems in back-end optimization.
    Most traditional place retrieval methods are based on single scan estimation~\cite{VPR:DBOW2} and strict outliers rejecting threshold is set to alleviate perceptual aliasing.
    However, many correct matches will also be rejected and therefore these methods only work for large overlaps.
    This work addresses this challenge by developing a hybrid loop closure estimation module.
    We notice that: 1) sequence matching \cite{2012SeqSLAM} can provide high recall and accuracy over long consistent overlaps but not in areas with few scan overlaps; 2) RANSAC-based single scan matching \cite{1981RANSAC, 2015Robust} can handle areas with short overlaps but may introduce inter-outliers (i.e., wrong matching between overlapped trajectories) for long-distance segments with similar geometry patterns.
    To detect matches with high accuracy, we formulate an adaptive loop closure detection mechanism by balancing the place retrieval mechanisms mentioned above.

    \bigskip
    \textbf{The contributions} of AutoMerge can be summarized as:
    \begin{itemize}
        \item AutoMerge provides a framework that can merge segments in city-scale environments without requiring initial coordinates estimations.
        Using this framework, we enable multi-agent map merging by being invariant to relative perspective differences and temporal differences.
        \item Within AutoMerge, we design an adaptive loop-closure detection module, which provides high recall and low false positive place retrievals, resulting in significantly reduced outliers in repeated environments during large-scale merging.
        \item AutoMerge can perform incremental map merging for single- and multi-agent systems. This procedure is invariant to the data streaming order from the different agents in the temporal and spatial domain and to the revisit times for the same area.
        \item Extensive evaluation on different large-scale datasets.
        We demonstrate detailed qualitative and quantitative analysis on the public \textit{KITTI}~\cite{DATASET:KITTI} dataset and on our city-scale and campus-scale datasets, which show that AutoMerge provides accurate map merging performance, and also that it has high generalization for unknown areas.
    \end{itemize}

    \bigskip
    \textbf{Novelty with respect to previous work~\cite{spherevlad,fusionvlad,adafusion}:}
    Our previous works investigate orientation-invariant~\cite{spherevlad} and translation-invariant~\cite{fusionvlad} 3D Loop Closure Detection (LCD). 
    Firstly, with a better understanding of features' robustness under different perspectives and our previous robust fusion-based data-association~\cite{adafusion}, AutoMerge presents an attention-enhanced multi-view fusion descriptor to improve the robustness under both translation and orientation differences simultaneously.
    Secondly, AutoMerge has an adaptive loop closure detection mechanism to maintain highly accurate loop closure detection with high recall rates.
    The above advantages allow AutoMerge to provide an offline map merging system for previously-stored 3D sub-maps, and an incremental map merging framework for single- and multi-agent mapping, which can further benefit the crowd-sourced mapping in current last-mile delivery and autonomous driving.

   \begin{table*} [ht]
        \caption{Comparison of different map merging approaches.}
        \centering
        \begin{tabular}{|C{3cm}|C{2cm}|C{3cm}|C{3cm}|C{2cm}|C{2cm}|}
            \hline
            \textbf{Method}
            & \textbf{Environments}
            & \textbf{Scale(km)}
            & \textbf{Single/Multi Robots}
            & \textbf{Offline}
            & \textbf{Online}
            \\
            \hline
            LAMP~\cite{Merge:LAMP} & Subterranean & $\leq 2$ & Multi & & \checkmark\\
            \hline
            Kimera-Multi~\cite{kimera_multi} & Outdoor & $\leq 2$ & Multi & & \checkmark\\
            \hline
            Multi-SLAM~\cite{multi_agent_fast_slam} & Indoor & $\leq 0.5$ & Multi & & \checkmark\\
            \hline
            RTAB-Map~\cite{rtabmap} & Indoor & $\leq 0.5$ & Single & & \checkmark\\
            \hline
            SegMap~\cite{segmap} & Outdoor & $\sim 10$ & Single & \checkmark & \\
            \hline
            AutoMerge (ours) & Outdoor & $\geq 100$ & Single/Multi & \checkmark & \checkmark\\
            \hline
        \end{tabular}
        \label{tab:CorrelationAveraging}
    \end{table*}

\section{Related Works}
\label{sec:related_works}

    Map merging is defined as reorganizing unordered sub-maps into one global and consistent map.
    Related works investigate map merging using different sensing modalities, including vision~\cite{kimera_multi}, sonar~\cite{multi_agent_sonar}, and LiDAR~\cite{Merge:LAMP}; LiDAR-based approaches have been widely applied in large-scale mapping due to their robustness to illumination changes and environmental conditions~\cite{VPR:SURVEY}.
    In this section, we mainly target the 3D map merging task, and investigate recent state-of-the-art approaches.
    We also briefly introduce the key techniques of 3D feature extraction and large-scale data association.

\subsection{Large-scale Map Merging}
\label{sec:map_merging}
    We review the literature on 3D SLAM systems for large-scale mapping, and refer the reader to~\cite{SLAM3D_survey} for a broader survey on 3D mapping.
    In general, 3D map merging is considered as map integration of different sub-maps with and without initial estimation.
    These 3D maps are typically represented as 3D point clouds, occupancy grids~\cite{octomap}, or 3D meshes~\cite{kimera}.
    Point cloud-based methods~\cite{Merge:LAMP}, mainly rely on geometric-based point cloud registration (such as ~\cite{icp,GICP}) to convert a point cloud into local 3D maps.
    The performance of the point-based approach is highly dependent on the robustness of 3D geometric features.
    In most cases, map merging algorithms operate using occupancy grids~\cite{multi_agent_fast_slam}, which are obtained by selecting a plane, e.g. a ground plane in the case of a wheeled robot.
    However, the simple representations in occupancy-based approaches cannot satisfy current requirements for long-term 3D navigation tasks.
    Kimera~\cite{kimera}, a mesh-based method, provides a deformation graph model to merge 3D meshes between different agents.
    This approach can ensure 3D mesh consistency when used in multi-agent distributed mapping~\cite{kimera_multi}.
    Most of the 3D map merging methods mentioned above are based on the assumptions that either all the sub-maps have the same initial pose~\cite{Merge:LAMP} or the mapping zones are restricted to a relatively small area~\cite{kimera_multi}.
    RTAB-Map \cite{rtabmap} is able to perform multi-session mapping using visual appearance-based LCD methods, through which a single robot can map separate areas in different sessions without giving relative initial poses between them.
    SegMap~\cite{segmap} can provide street block-like global map merging, but its data association is highly reliant on the segmentation of distinguishable semantic objects, which is hard to satisfy in a city-scale or campus-scale map merging task.
    In all the above methods, the success of large-scale map merging is highly reliant on accurate data association between different segments.
    However, accurate place feature extraction and data association are difficult to guarantee, especially for large-scale map merging. 
    In Table.~\ref{tab:CorrelationAveraging}, we compare AutoMerge with existing merging methods; our method is able to merge large-scale maps for single- and multi-agent scenarios, both offline and online.
    
\subsection{Place Feature Extraction}
\label{sec:place_feature}
    Place feature extraction is the core module for providing accurate data association for map merging.
    LiDAR is the preferred sensor used for place feature extraction, since LiDAR inputs are inherently invariant to illumination changes.
    A representative example of the point-based approach to place feature extraction is PointNetVLAD~\cite{PR:pointnetvlad}; in this work, Mikaela~\textit{et al} utilized PointNet~\cite{feature:pointnet} to extract local features and cluster them into a global place descriptor via the deep vector of locally aggregated descriptors (VLAD)~\cite{NetVLAD}. 
    This work enables learning the 3D place features directly from a point cloud.
    But PointNetVLAD omits the inner connections between points, which will significantly reduce the localization accuracy under different viewpoints.
    Based on the extended 3D point feature extraction of PointNet++~\cite{pointnet++}, LPD-Net~\cite{liu2019lpd} takes both point cloud and handcrafted features as inputs and introduces a graph-based aggregation module to learn multi-scale spatial information.
    \cite{LPR:minkloc3d}, \cite{LPR:minkloc3dv2} apply Feature Pyramid Network~\cite{Feature:PFN} to extract local features based on the sparse voxelized representation.
    \cite{LPR:SVT-Net} utilizes a transformer module on top of the 3D sparse convolution network to learn the long-range dependencies. \cite{LPR:PPT-Net} employs a pyramid transformer module to extract the local features at different resolutions in order to further explore the spatial contextual information.
    \cite{LPR:Rank-Point} proposes an efficient strategy based on visual consistency to evaluate the registration between the query frame and frames in the initial retrieval list.
    On the other hand, Projection-based approach has also been widely applied in non-learning and learning-based methods.
    Non-learning based methods, such as distance and angle based features ESF~\cite{esf}, structure based Scan-Context~\cite{cohen2018spherical,lpr:scan++}, and histogram-based features SHOT~\cite{shot}, have shown accurate recognition ability in city-scale environments, but are sensitive to large translation difference.
    \cite{lpr:stable} utilizes adversarial feature learning to improve the generalization ability of projection methods for local translation differences.
    In our previous work SphereVLAD~\cite{seqspherevlad}, we use spherical harmonics to obtain a viewpoint-invariant descriptor for 3D place recognition.
    In OverlapNet~\cite{PR:overlapnet}, the authors utilize a deep neural network to exploit different cues from LiDAR to estimate loop closures and the relative orientations.
    Hui~\textit{et al.}~\cite{lpr:Point-Transformer} introduced a pyramid point cloud transformer network, based on the recent development of attention networks~\cite{attention_all_your_need}; this work improves the place recognition ability for PointNetVLAD. Ma~\textit{et al.}~\cite{lpr:overlaptransformer} extend the loop closure detection ability of OverlapNet with an additional transformer module.
    In our previous work, FusionVLAD~\cite{fusionvlad}, we proposed a deep fusion network integrating different perspectives to learn features that are resistant to translation/orientation differences.

\subsection{Large-scale Data Association}
\label{sec:data_association}
    Data association is critical for estimating the single and inter-connections between sub-maps in the multi-agent map merging task.
    In current SLAM frameworks~\cite{SLAM3D_survey}, robots utilize a combination of global descriptors (e.g., bag-of-words vectors~\cite{VPR:DBOW2} and learned full-image descriptors~\cite{VLAD}) to find the overlaps between different sub-maps. 
    However, single scans may include measurement noise, especially in large-scale city environments~\cite{DATASET:KITTI}, which can cause incorrect matches between different segments.
    SeqSLAM~\cite{2012SeqSLAM} provides a sequence-based place recognition method, which can improve recognition accuracy using the difference of residuals of a sequence of observations under changing environmental conditions.
    In our previous work~\cite{spherevlad}, we integrated a sequence-matching method with our SphereVLAD to provide viewpoint-invariant place recognition ability under changing 3D environments.
    Recently, Shan~\textit{et al.}~\cite{lpr:ransac} provided a RANSAC-based data association method to remove outliers in data association.
    However, most of the above methods are focused on single-agent inner data association, but few show promising results for large-scale multi-agent map merging tasks, where the may exist significant perspective and appearance differences between observations.

\begin{figure*}[t]
	\centering
    \includegraphics[width=\linewidth]{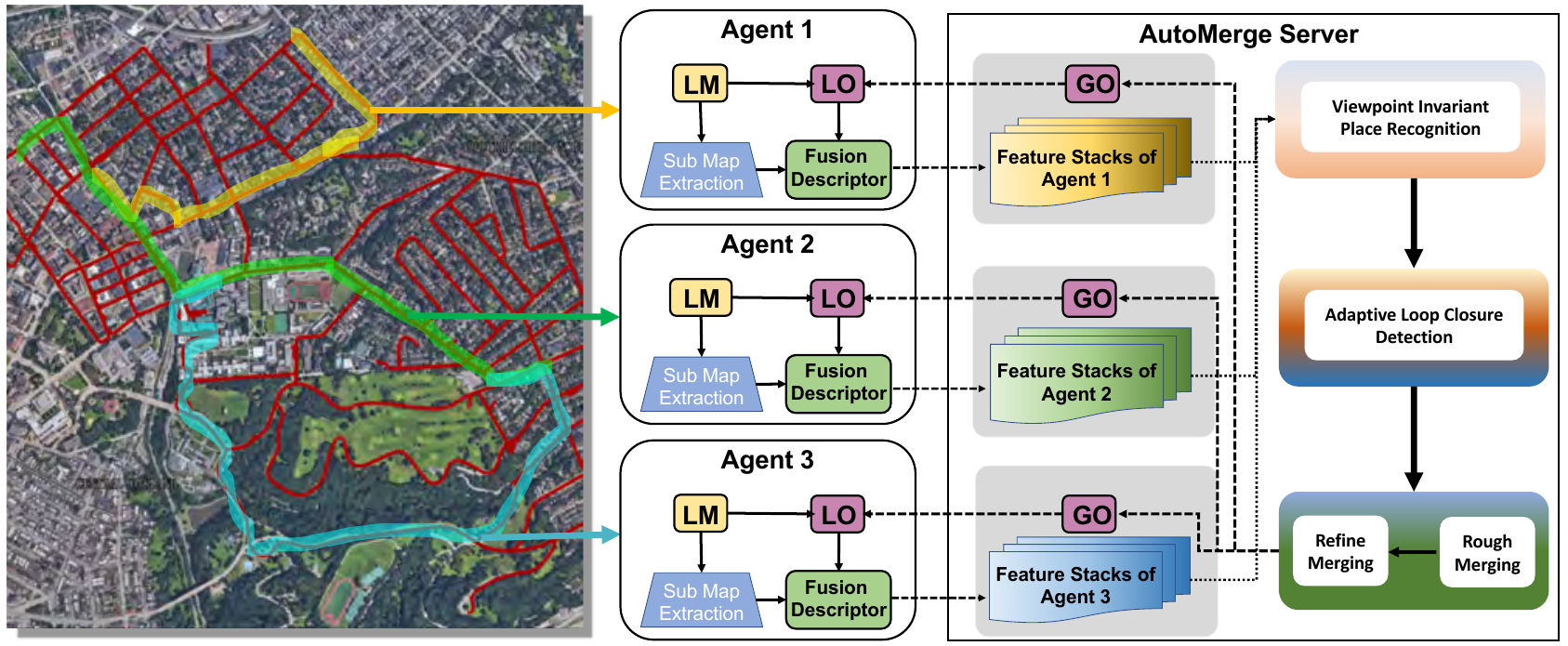}
	\caption{\textbf{AutoMerge system framework.}
    {AutoMerge supports offline global map merging tasks.
    In the offline mode, AutoMerge can use the previously-stored submaps for direct global data association and map merging.
    In the online mode, given LiDAR odometry estimates, each agent can extract adaptive place descriptors from local sub-maps and stream them back to the AutoMerge Server.
    Due to  the viewpoint-invariance of these descriptors, AutoMerge can estimate accurate data associations between different segments.
    The sub-maps are merged into a global map using a rough global optimization method (\textbf{GO}), and each agent can estimate in parallel their global location through a local optimization (\textbf{LO}) method.}
    }
	\label{fig:pipeline}
\end{figure*}

\section{SYSTEM OVERVIEW}
As shown in Fig.~\ref{fig:pipeline}, AutoMerge provides an automatic map merging system for the large-scale single-/multi- agent mapping tasks.
Each agent is equipped with a LiDAR mapping module to enable the self-maintained sub-map generation and odometry estimation.
The AutoMerge system consists of three modules: 1) fusion-enhanced place descriptor extraction, 2) an adaptive data-association mechanism to provide high accuracy and recall for segment-wise place retrievals, and 3) a partially decentralized system to provide centralized map merging and single agent self-localization in the world frame.

\textbf{Problem Formulation}
\label{sec:formulation}
We define the trajectory list as $V_N=\{v_1,v_2,...,v_n\}$, each agent starts from the random position in the unknown map without initial coordinate knowledge and uploads the local odometry and 3D observations to the server incrementally.
And please note that the data streaming order of $V_N$ is random and in an incremental manner.
We assume can estimate a confidence score $\omega_{ij}\sim [0,1]$ based on the overlap between two trajectories $v_i$ and $v_j$.
The task of AutoMerge is to generate the accurate global maps $M_{global}$ combined by locally connected sub-groups $A_M=\{A_1,...,A_m\}$, based on the data stream from the trajectory list $V_N$ and their relative confidence matrix $\Omega_{N\times N}=\{\omega_{11},..., \omega_{1n};...;\omega_{n1},...,\omega_{NN}\}$, ignoring the order and completeness of the reserved data.

\textbf{Fusion-enhanced Descriptor:}
\label{sec:decent_local}
Each agent runs the decentralized mapping sub-system as an extension of LiDAR-inertial odometry estimation~\cite{LOAM:zhang2014loam}, which also provides a sub-map extraction module for onboard adaptive descriptor extraction.
Such a descriptor has the following advantages:
1) it is translation-invariant due to the local translation-equivalent property of 3D point-clouds~\cite{adafusion}, 2) it is orientation-invariant due to the rotation-equivalent property of spherical harmonics~\cite{spherevlad}, and 3) it is light-weight compared to the original raw sub-maps.
Thus, a single agent can provide paired viewpoint-invariant place descriptors and ego-motion to the AutoMerge server system through lower bandwidth communication.

\textbf{Adaptive Loop Closure Detection:}
Spurious loop closures are frequent in environments with repetitive appearances, such as long streets.
On the one hand, false positive place retrievals may easily break the global optimization system, and ideally $100\%$ accuracy can avoid these optimization failures for large-scale mapping.
On the other hand, low recalls can provide partial data association, which will affect global optimization performance.
Hybrid loop closure detection takes advantage of sequence matching to provide continuous true positive retrievals over long overlaps, and RANSAC-based single frame detection for local overlaps. 
By analyzing the feature correlation between segments, we can balance the place retrievals from sequence-/single- frame matching to provide accurate retrievals for offline/online LCD.

\textbf{Incremental Merging:}
Traditional centralized map merging~\cite{Merge:LAMP} is usually reliant on initial relative odometry estimation and geometry-based point cloud registration.
In contrast, AutoMerge uses the paired place descriptors and ego motions of each agent to capitalize on loop closure opportunities (high accuracy and recall) for correction, despite a large amount of odometric drift.
Using this hybrid loop closure detection method, AutoMerge performs a rough centralized global map optimization.
Given the obtained information $V_N$ and their relative confidence matrix $\Omega_{N\times N}$, AutoMerge utilizes the spectral clustering method to adaptively merge different trajectories into different sub-groups $A_M$.
And our system can make each individual sub-group is well connected for the trajectories within, and no wrong matches are built between different sub-groups.
This mechanism ensure the incremental map merging ability, when the data stream of different agent in the list $V_N$ come with different time-order or completeness.

\begin{figure*}[t]
    \centering
        \includegraphics[width=\linewidth]{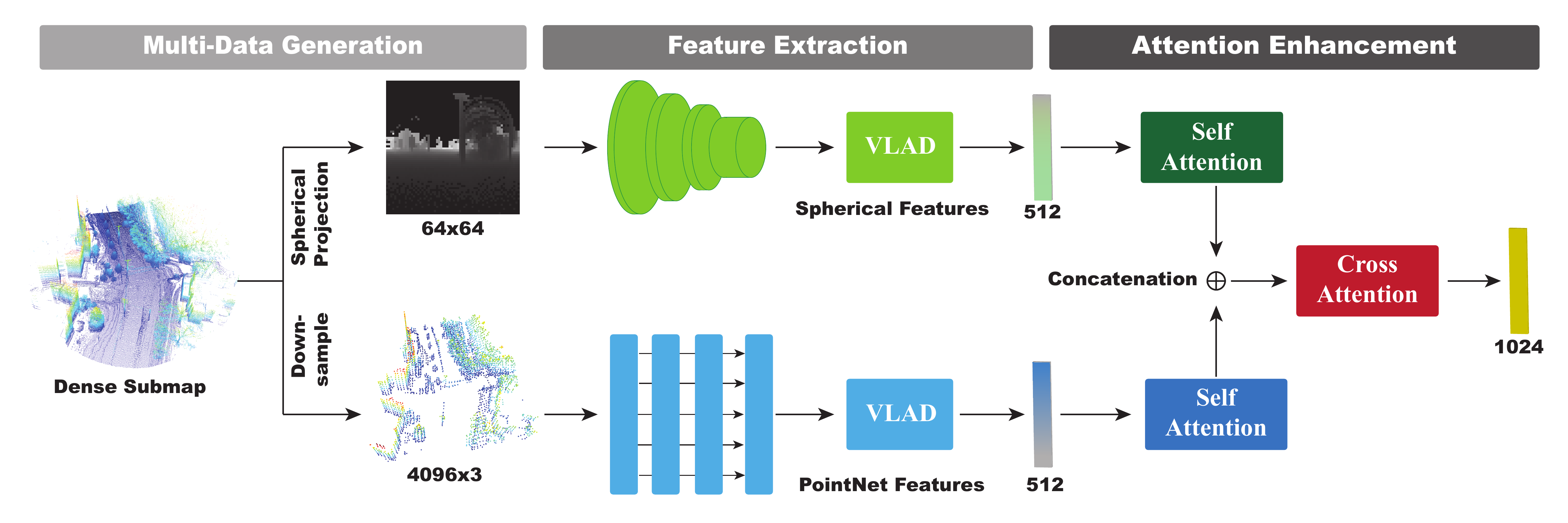}
        \caption{\textbf{The network structure of the AutoMerge descriptor extraction.}
        To provide viewpoint-invariant descriptor extraction, our network includes a point-based branch to improve the robustness of translation differences and a sphere-based branch for rotation differences.
        Finally, AutoMerge utilizes a deep fusion mechanism between the two branches, which can enhance the individual branches and the joint branch simultaneously.
        }
    \label{fig:flowchart}
\end{figure*}

\section{Fusion-enhanced Descriptor Extraction}
\label{sec:fuse_dis}
    As analyzed in Sec.~\ref{sec:related_works}, point-based approaches show better performance against translation differences when compared with projection-based methods, whereas projection-based methods show better accuracy against orientation differences.
    Our fusion-enhanced descriptor balances the advantages of both point- and projection-based approaches with a multi-perspective feature extraction network.
    As shown in Fig.~\ref{fig:flowchart}, the network includes two core components: 1) a multi-perspective feature extraction module, 
    and 2) an attention place feature fusion module.
    
\subsection{Multi-perspectives Feature Extraction}
    \label{sec:MFE}
    Due to the sparsity and occlusion problems of a raw LiDAR scan, a 3D observation will vary when gathered under different viewpoints.
    To provide stable multi-perspective feature extraction, we first accumulate the point cloud into local dense maps.
    This mechanism can provide a consistent local dense map based on LiDAR odometry estimation, which has been explained in detail in our previous work~\cite{fusionvlad}.
    In the following subsections, we provide details on how we utilize point-based and projection-based feature extraction in our method.
    
    \bigskip
    \noindent \textbf{Point-based Feature Extraction:}
    \label{sec:point-view}
    In this branch, we adopt the idea of PointNetVLAD\cite{PR:pointnetvlad} to extract the global descriptor. 
    Given a local dense map, we query the points set $P=\{p_1,...p_N|p_n \in \mathbb{R}^3\}$ within a $80m \times 80m$ bounding box and preprocess it as shown in \cite{PR:pointnetvlad}. 
    Then, $P$ is fed into PointNet\cite{feature:pointnet} to extract local features $F_p=\{f_1,...f_N|f_n \in \mathbb{R}^C_{p}\}$.
    With the help of a NetVLAD layer\cite{NetVLAD}, the global descriptor $V_{point}$ is obtained by aggregating local point features. 
    Finally, the global descriptor $V_{point}$ is run through  a fully connected layer to yield a compact descriptor.

    \bigskip
    \noindent \textbf{Projection-based Feature Extraction:}
    \label{sec:sph-view}
    To obtain viewpoint invariance, we utilize the spherical convolution\cite{cohen2018spherical} to extract local features from a spherical projection of the point cloud. 
    Using the local dense maps, we query the points within a range of $50m$ and project them into a panorama using the method mentioned in \cite{fusionvlad}. 
    Then the corresponding spherical projection $SP \in \mathbb{R}^{H \times W}$ is fed into 4 layers of spherical convolutions to generate local features $F_s \in \mathbb{R}^{C_s \times \alpha \times \beta \times \gamma}$ which contain features sampled from angles in all three axes.
    A NetVLAD~\cite{NetVLAD} layer is used to find the spatial similarities between local features and reorder them in a specific manner.
    Finally, the global descriptor $V_{sphere}$ is also run through a fully connected layer to reduce the feature dimensions.

    \begin{figure}[t]
    	\centering
        	\includegraphics[width=\linewidth]{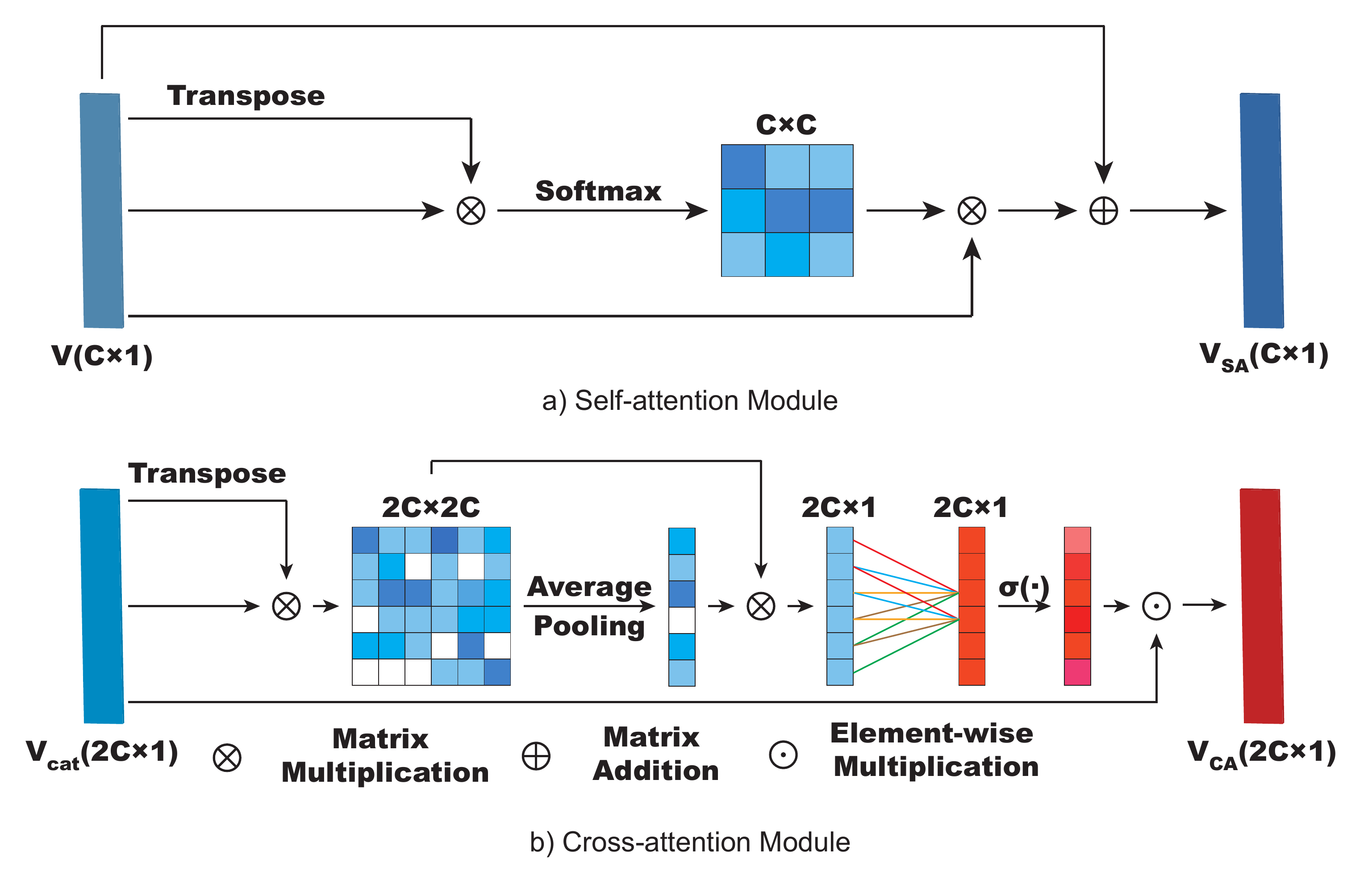}
    	\caption{\textbf{Attention-enhanced Feature Extraction.}
    	For each branch, AutoMerge applies a self-attention layer to improve the network feature extraction.
    	Between the two branches, AutoMerge also applies a cross-attention layer to enhance their inner connections.
    	}
    	\label{fig:fusion_layer}
    \end{figure}

\subsection{Attention Fusion}
\label{sec:fusion_network}
Our attention fusion module consists of two self-attention modules providing contextual information for $V_{point}$ and $V_{sphere}$ individually, and a cross attention module which aims to reweigh the importance of channels within the concatenation of $V_{point}$ and $V_{sphere}$.

    \bigskip
    \noindent \textbf{Self-attention Feature Enhancement:}
    Each channel of the global descriptor can be interpreted as a specific response and different combinations of channels can be regarded as different patterns of the environment \cite{att:dual_attention}. 
    However, when extracting the local features, PointNet\cite{feature:pointnet} only considers each point independently and the receptive field of the spherical convolution is also limited by the number of layers, which leads to a lack of inter-channel dependencies in global descriptors.
    By exploring the inter-dependencies between channels, we can therefore enhance the semantic information representation of the global descriptor.
    \begin{align}
        V_{SA} &= V + \gamma \text{SoftMax}(V V^{T})^{T}V
    \end{align}
   
       \begin{figure*}[ht]
    	\centering
        	\includegraphics[width=\linewidth]{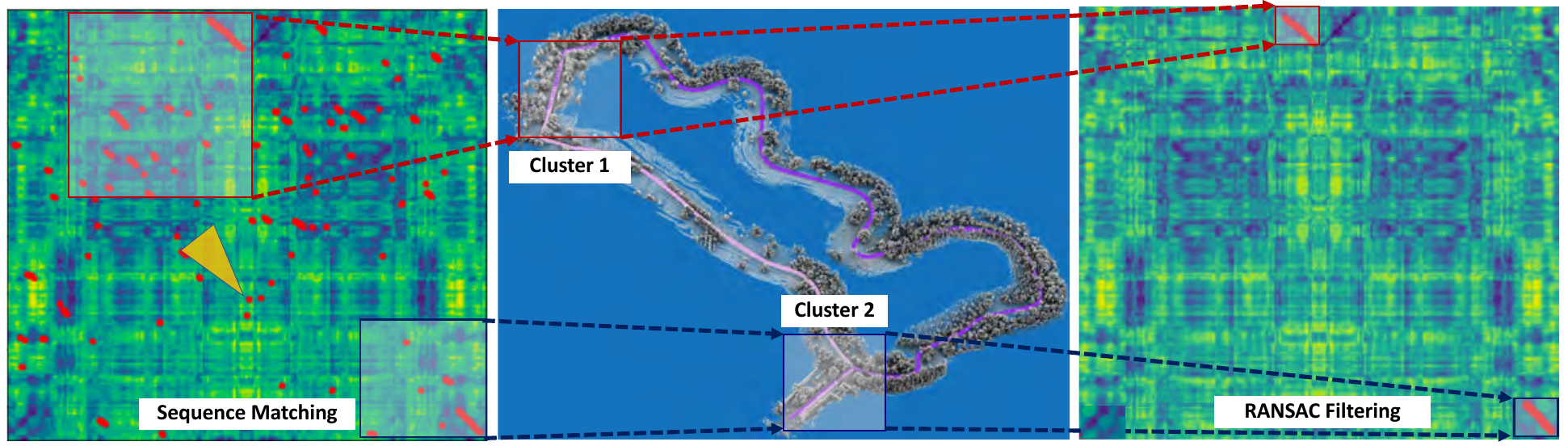}
    	\caption{\textbf{Illustration of Adaptive Loop Closure Detection.}
    	(i) The figure shows the difference matrix between two segments, where the red points indicate the potential matches via sequence matching~\cite{2012SeqSLAM}. 
    	(ii) The matches are clustered into individual zones via K-means based on their feature distances.
    	(iii) RANSAC is utilized to exclude outliers that can not satisfy Euclidean constraints.
    	}
    	\label{fig:adaptive_lcd}
    \end{figure*}
    
    Given a global descriptor $V \in \mathbb{R}^{C \times 1}$, we obtained the attention map $A \in \mathbb{R}^{C \times C}$ by directly multiplying $V$ and its transpose and applying a SoftMax function on the result along the row direction.
    Each element $A_{ji}$ represents the impact of $j^{th}$ channel on $i^{th}$ channel.
    Then, we multiply the transpose of attention map $A$ with $V$ to generate a weighted sum of every channel which contains the inter-channel dependencies.
    Finally, we multiply the result with a learnable parameter $\gamma$ to scale the inter-channel dependencies and add it with $V$.

    \bigskip
    \noindent \textbf{Cross-attention Feature Reweighing:}
    While inter-channel dependencies can provide contextual information, they can also evaluate the contribution of each channel.
    During feature fusion, there are situations that are only beneficial to one of the branches, and so simple concatenation of the two global descriptors will lead to a large performance decrease.
    Inspired by~\cite{att:channel_attention}, our cross-attention feature reweighing module learns the inter-channel dependencies and emphasizes the more meaningful channels and neglects the irrelevant channels.
    
    The network structure is illustrated in Fig.\ref{fig:fusion_layer}. The input is concatenation of two global descriptor $[V_{point}, V_{sphere}]$ denoted as $V_{cat} \in\mathbb{R}^{2C \times 1}$.
    We directly get the correlation matrix $E$ from multiplication of $V_{cat}$ and its transpose.
    \begin{align}
        E = V_{cat} V_{cat}^{T}
    \end{align}
    Then, we average the elements in each row of the $E$ to aggregate the responses of each channel, and multiply $E$ with the result to obtain channel correlation weight $\alpha_{corr}$.
    We further utilize a fully connected layer to exploit the dependencies of channels and apply a Sigmoid function to narrow the channel importance weight $\alpha_{w}$ within $[0,1]$.
    \begin{align}
        \alpha_{corr} = E\otimes \text{Ave\_Pool}(E) \\
        \alpha_{w} = \text{Sigmoid}(W(\alpha_{corr}))
    \end{align}
    Finally, we apply an element-wise multiplication between channel importance weight $\alpha_{w}$ and $V_{cat}$ to yield the attention-reweighted global descriptor.
    \begin{align}
        V_{CA} = \alpha_{w} \odot V_{cat}
    \end{align}

\subsection{Learning Metrics}
\label{sec:loss}
To enable end-to-end training for our network, we utilize the "Lazy quadruplet" loss metric.
Sets of training tuples are selected from the training dataset and each of these training tuples is composed of four components: $\mathcal{S} = [S_a, \{S_{pos}\}, \{S_{neg}\}, S_{neg^{*}}]$, where the $S_a$ represents the query frame location at the ground truth position, $\{S_{pos}\}$ stands for a set of ``positive" frames whose distance to $S_a$ is less than the threshold $D_{pos}$, $\{S_{neg}\}$ denotes a set of ``negative" frames whose distance to $S_a$ is strictly larger than threshold $D_{neg}$ and $S_{neg^{*}}$ represents a frame whose distance to $\{S_{neg}\}$ is strictly larger than $D_{neg}$.
In our case, these two thresholds are set as $D_{pos}=10m$ and $D_{neg}=50m$.
The lazy quadruplet loss is defined as
\begin{align}
    L_{lazyQuad}(\mathcal{S}) = &\max_{i, j}([\gamma + \delta_{pos_i} - \delta_{neg_j}]_{+})  +  \\ &\max_{i,k}([\alpha + \delta_{pos_i} - \delta_{neg_k^{*}}]_{+}) \nonumber
\end{align}
where $\alpha$ and $\beta$ are the constant threshold giving the margin and $[...]_{+}$ denotes the hinge loss. 

\section{Adaptive Loop Closure Detection}
\label{sec:lcd}
    In our AutoMerge framework, loop closures are required to be accurate (few false positives) and robust (high recall), though these two properties are contradictory.
    As shown in Fig.~\ref{fig:adaptive_lcd}, our adaptive loop closure detection module can estimate the stable data-association while ignoring the potential outliers, through our sequence matching and RANSAC filtering mechanisms.
    In this section, we will introduce details.

\subsection{Adaptive Candidates Association}
    In order to find possible loops among multiple segments, they are grouped into pairs for every two segments $T_i$ and $T_j$ $(i \ne j)$. Each $T_i$ has point cloud sub-maps separated by a constant distance and the corresponding poses $\mathbf{T}_i = \{\mathbf{T}_i^k\}$, both of which are obtained from the odometry. These sub-maps are then encoded with our fusion-enhanced descriptor and represented as feature $\boldsymbol{f}_i = \{\boldsymbol{f}_i^k\}$. 
    The similarity of places (i.e. sub-maps) of different segments can be revealed in the difference matrix $\mathbf{D} = \mathrm{d}(\boldsymbol{f}_i, \boldsymbol{f}_j) \in \mathbb{R}^{N_i \times N_j}$, where $\mathrm{d}(\cdot)$ is the cosine distance and $N_i$ and $N_j$ are the number of sub-maps in $T_i$ and $T_j$ respectively.
    
    Our adaptive LCD method works on loop candidates $\mathcal{C} = \{(k_i, k_j)\}$ where $k_i$ and $k_j$ are the index of submaps in $T_i$ and $T_j$, showing the association of places in a segment pair.
    We acquire $\mathcal{C}_\mathrm{seq}$ by applying sequence matching \cite{2012SeqSLAM} on the difference matrix $\mathbf{D}$.
    However, as we can see in Fig.~\ref{fig:adaptive_lcd}, the raw match result still exists lots outliers.
    To filter them out, we utilize kmeans to cluster the potential matches into different zones ${\mathcal{C}_\mathrm{seq}}_i,i=1,...,k$ via k-means based on the feature distances.
    Then for $i$-th zone, we adopt the idea of RANSAC to select correspondences from ${\mathcal{C}_\mathrm{seq}}_i$, with an edge-based geometric consistency principle to check the correctness of the proposal of correspondences.
    Specifically, within each iteration, the relation
    \begin{equation}
        \|\mathrm{edge}_i \|_2 \ge \beta \|\mathrm{edge}_j \|_2, \qquad
        \|\mathrm{edge}_j \|_2 \ge \beta \|\mathrm{edge}_i \|_2
    \end{equation}
    is checked between $n$ samples $(k_i, k_j)$, where the edges are formed by every two samples $(k_i^1, k_j^1)$ and $(k_i^2, k_j^2)$, and $\beta \in [0, 1]$ controls the degree on equality of edge length.
    Through the above mechanism, AutoMerge can filter out most outliers.
    

\color{red}

\color{black}

\section{Incremental Merging}
\label{sec:inc_merge}
For the large-scale merging task, we may encounter a case where there exist more than two groups of segments with overlaps.
The overlaps between segments are limited at the early stage and can be extended at the late stage.
In merging, dividing all segments into groups with stable connections is essential for incremental factor graph optimization.
This section will introduce the details of our incremental merging mechanism.

\subsection{Multi-agent Clustering}
\label{sec:spectral}

Firstly, we formulate the incremental merging task into a traditional spectral clustering problem~\cite{spectral_cluster}.
Assume there exists $V=\{v_1, ..., v_n\}$ agents running independently, and we define the inner connection $\omega_{ij}\sim [0,1]$ between $v_i$ and $v_j$, which indicates the overlap confidence of agent $v_i$ and $v_j$ existing stable overlaps.
Without losing generality, we define a weighted graph $G=(V,E)$. $E$ represents the edge connections, which satisfies $\omega_{ij}=\omega_{ji}$.
We define $A_i,i=1,...k$ as the subset of $V$, and satisfies $A_1 \cup ... \cup A_k = V, A_i\cap A_j=\emptyset$, $\bar{A_i}$ is the complement of $A_i$.
And $W(A_i, \bar{A_i})$ is weighted adjacency matrix, which is defined as,
\begin{align}
    W(A_i, \bar{A_i}) := \sum_{k \in A_i, l \in \bar{A_i}} \omega_{kl}
    \label{eq:adjacency}
\end{align}

\begin{figure}[t]
	\centering
    	\includegraphics[width=\linewidth]{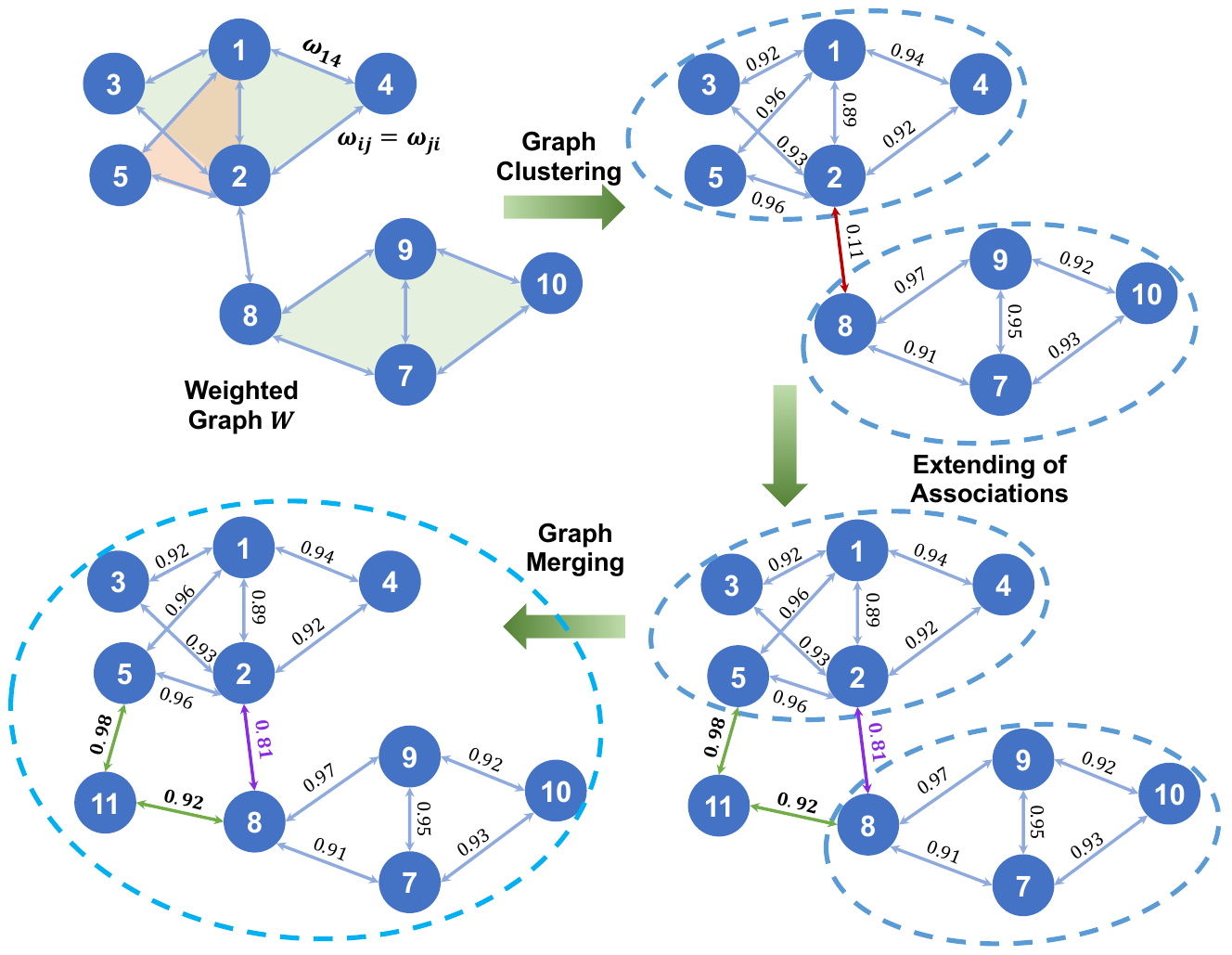}
	\caption{\textbf{Incremental Merging.}
	(1) Graph $G=(V,E)$ is constructed with $V=\{v_1,...,v_n\}$ agents and their connections $\omega_{ij}\in E, i\neq j$.
	(2) AutoMerge can incrementally merge agents into different sub-groups by maximizing the inner connections within each sub-group and minimizing the connections between sub-groups.
	(3) When a new node $v_{11}$ builds new confident connections with $v_5$ and $v_8$, or previous connection $\omega_{2,8}$ is re-enhanced, AutoMerge can update the sub-groups into a joint group.
	}
	\label{fig:inc_merge}
\end{figure}

From the loop closure detection perspective, the inner connection $\omega_{ij}$ between agent $v_i$ and $v_j$ is based on overlap length and place recognition quality.
Thus, we define inner connection $\omega_{ij}$ as,
\begin{align}
    \omega_{ij} = \left\{ 
    \begin{array}{cc}
         &  \exp \left(-\frac{\|F_i - F_j\|_2^2+C_{\omega}}{2L_{ij}^2+\epsilon} \right), i\neq j \\
         & 0, i=j
    \end{array}
    \right.
    \label{eq:connection}
\end{align}
where $F_i$ indicates the extracted overlap place features from agent $v_i$, and $L_{ij}$ is the length of overlap area.
$C_{\omega}$ is a hyper-parameter to control the $\omega_{ij}$'s dependence on the overlaps' length, and $\epsilon=1e-4$ is a constant parameter.
In the extreme cases where $\|F_i - F_j\|_2^2 \ll C_{\omega}$, the weighting $\omega_{ij}\sim \exp \left(-\frac{C_{\omega}}{2L_{ij}^2} \right)$ will mainly depend on the length of overlaps.
Based on the above equation, we can also define degree matrix $D$, where $d_{ii}=\sum_{j=1}^n \omega_{ij}$ and is the connection measurement between agent $v_i$ with all other agents $\bar{v_i}$.
According to spectral clustering~\cite{spectral_cluster}, the incremental merging task can be defined as a mincut problem, 
\begin{align}
    \min cut(A_1,...,A_k) := \min \frac{1}{2}\sum_{i=1}^k W(A_i, \bar{A_i})
\end{align}
The major limitation of the above mincut is that it will simply separate one individual agent $v_i$ from the rest of agents $\bar{v_i}$, which is not our desired map segmentation. 
To maintain sub-groups with a large size, we utilize the object function from Ncut~\cite{ncut},
\begin{align}
    \min Ncut(A_1,...,A_k) &:= \min \frac{1}{2}\sum_{i=1}^k \frac{W(A_i, \bar{A_i})}{vol(A_i)} \\
    vol(A_i) &:= \sum_{j\in A_i} d_{jj}
\end{align}
where $vol(A_i)$ is the measurement of inner connections among the sub-group $A_i$.
From the power consumption perspective, incremental clustering is trying to find the best segment option with a minimum penalty to divide the original agents into different consistent sub-groups.

The solution to the Ncut problem is detailed in the reference~\cite{spectral_cluster}, and the standard spectral cluster approach is shown in Algorithm~\ref{alg:spectral}. 
Given the agent list $V=\{v_1,...,v_n\}$, we can calculate the similarity matrix (Eq.~\ref{eq:connection}), degree matrix $D$, and corresponding Laplace matrix $L$.
The eigenvalues $\lambda_k, k=1,...,n$ can indicate the clustering status.
In theory~\cite{spectral_cluster}, if there exist $k$ different sub-groups $\{A_1,...,A_k\}$ without connections $W(A_i, A_j)=0, i\neq j$, the number of eigenvalues $\lambda_i = 0$ equals to $k$.
In the map merging problem, partial overlaps between different sub-groups may exist, thus we set a control threshold ($\lambda_{max} \leq \theta$) to estimate the best sub-groups size.
Based on the first $k$-dimension of eigenvectors $\mathcal{U}$, we can construct a key matrix $\mathcal{K}^{n\times k}$, and cluster to $k$ classes though $k$-means.
Through the above operation, AutoMerge can cluster agents into $k$ sub-groups.

\begin{algorithm}[t]
\caption{Incremental Clustering}
\label{alg:spectral}
\begin{algorithmic}[2]
    \renewcommand{\algorithmicrequire}{\textbf{Input:}}
    \renewcommand{\algorithmicensure}{\textbf{Output:}}
    \Require Agent list $V=\{v_1,...,v_n\}$
    \Ensure Clusters $A_1,..., A_k$ with $A_i \in \{j|y_j \in \mathcal{R}^k\}$
    \State Construct similarity graph $W$, and $\omega_{ij}$ based Eq.~\ref{eq:connection} 
    \State Construct degree matrix $D$, and $d_{ii}=\sum_{j=1}^n \omega_{ij}$
    \State Calculate Laplace matrix $L=W-D$
    \State Compute the eigenvectors $\mathcal{U}=\{u_1,...,u_n\}$ and eigenvalues $\{\lambda_1,...,\lambda_n\}$ from $L$
    \State Sort eigenvectors $\mathcal{U}$ based on eigenvalues 
    \State Determine the cluster numbers $k$ based on $\lambda_{1,...,k}\leq \theta$
    \State Construct key matrix $\mathcal{K}=\{u_1,...,u_k\}$, and get $y_i \in \mathcal{R}^k$ from $i$-th row from $\mathcal{K}$
    \State Cluster points $(y_i)_{i=1,...,n}$ into $k$ clusters with $k$-means.
\end{algorithmic}
\end{algorithm}

\subsection{Incremental Merging}
\label{sec:merging}
Recall that in Figure.~\ref{fig:pipeline}, the data from different agents will stream to the AutoMerge server in random order.
To achieve stable and accurate incremental merging, the AutoMerge operations contain the following three steps:
\begin{itemize}
    \item \textbf{Step1:} When each agent $v_i$ streams their observations and local place descriptors back to the server, AutoMerge will automatically detect the potential overlaps based on our Adaptive loop closure detection mechanism as stated in Section.~\ref{sec:lcd},
    and parallel estimate the overlaps' transformation via the method mentioned in \cite{sorkine2017least}.
    \item \textbf{Step2:} As shown in Fig.~\ref{fig:inc_merge}, when new observation for agent $v_{i}$ received, AutoMerge will automatically estimate corresponding weightings $\omega_{ij}, i\neq j$ between $v_{i}$ and $v_j\in \bar{v}_i$; 
    when new overlaps are observed for existing agents, the previous weak connection ($\omega_{2,8}$) is further enhanced.
    \item \textbf{Step3:} Given the received agent lists $V_N=\{v_1,...,v_{N}\}$ and their relative overlap weighting $\omega_{ij}$, the system applies the graph clustering based on Section.~\ref{sec:spectral} to generated individual stable sub-groups $A_M=\{A_1,..., A_m\}$.
    \item \textbf{Step4:} Based on updated graphs, AutoMerge applies the standard back-end pose graph optimization(GTSAM~\cite{gtsam}) for each sub-graph in $A_M$.
    The optimized position is sent back to all the agents for global pose estimation. Then go back to \textbf{Step1}.
\end{itemize}
In the above operations, the core of AutoMerge merging is triggered by Step2 and Step3 especially, which can adaptively fuse new observations into the global mapping ignoring their relative data streaming order.
Therefore, AutoMerge can transform the current offline high-resolution mapping into an incremental version.


    \begin{figure*}[t]
        \centering
        \includegraphics[width=\linewidth]{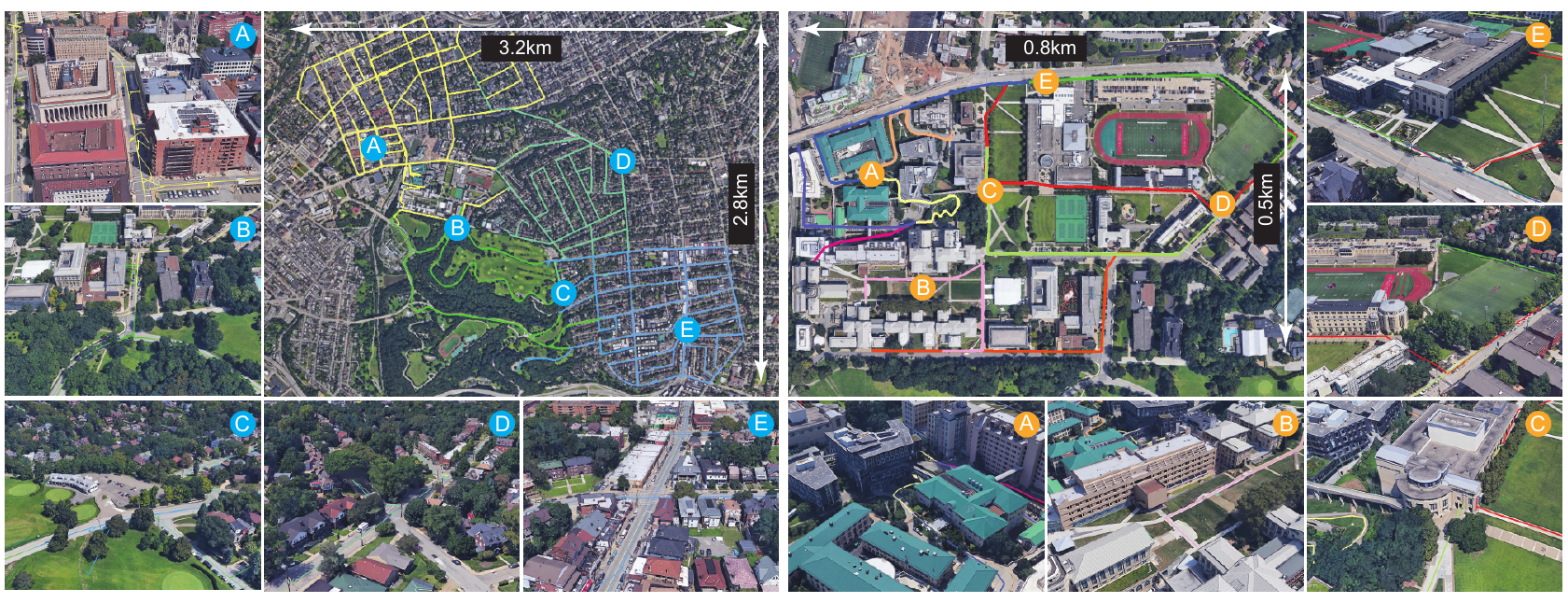}
        \caption{\textbf{Self-collected Datasets.}
        Both \textit{Pittsburgh} and \textit{Campus} datasets are collected with our data-collection device (Velodyne-16 and Xsens MTI-300 IMU). 
        The \textit{Pittsburgh} dataset includes 4 zones (colored in yellow, green, cyan, and blue), which covers street blocks, residential areas, parks and commercial buildings. 
        The \textit{Campus} dataset is shown in the pictures on the right, which covers the main campus area of Carnegie Mellon University.
        }
        \label{fig:dataset}
    \end{figure*}
    
    \begin{figure}[t]
    	\centering
    	\includegraphics[width=\linewidth]{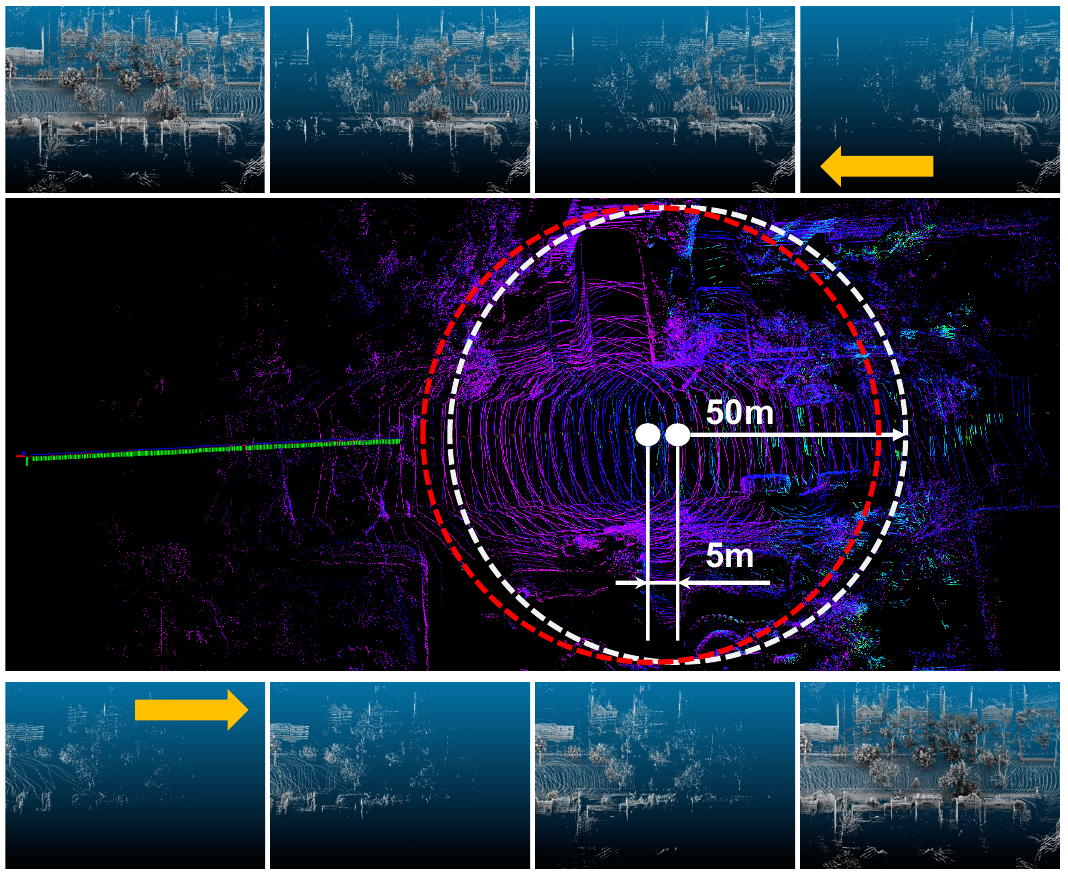}
    	\caption{\textbf{Sub-map Generation.}
    	The dense maps are generated by accumulating the LiDAR scans into local sub-maps, which have a $50m$ radius and $5m$ offset from their neighbors.
        For either a forward or reverse traversal direction, the generated sub-maps for the same area share the similar geometric structures.
    	}
    	\label{fig:dense_map}
    \end{figure}

\section{Datasets and Criteria}
\label{sec:data_crit}
    To evaluate the map merging accuracy, we choose the well-known \textit{KITTI}~\cite{DATASET:KITTI} dataset, one city-scale dataset collected in the City of Pittsburgh with around $\mathbf{120}$ km of trajectories in total, one campus-scale dataset collected within Carnegie Mellon University with $\mathbf{4.5\times 8}$ km trajectories.
    The last two datasets are self-recorded with our data-collection platform, and they contain multiple revisits, as well as translation and orientation differences.
    In this section, we describe the datasets, target methods, and evaluation criteria.
    
    \bigskip
    \noindent \textbf{Merging Datasets:} To cover various scenarios in our datasets, we travel through different types of areas over our self-gathered datasets, and we include multiple revisits.
    The detailed characteristics of each dataset and the environment will be provided in the following descriptions.
    Fig.~\ref{fig:dataset} shows the overlaid segments on an aerial map, which illustrates the segment shapes, scales, and areas. The details are summarized in Table~\ref{tab:dataset}.
    
   \begin{table} [htbp]
        \caption{Comparison of different map merging approaches.}
        \centering
        \begin{tabular}{|C{1.5cm}|C{4cm}|C{1.5cm}|}
            \hline
            \textbf{Dataset}
            & \textbf{Environments}
            & \textbf{Scales (km)}
            \\
            \hline
            KITTI~\cite{DATASET:KITTI} & Street & $15\times 1$\\
            \hline
            Pittsburgh & Street, Residential, Terrain & $120\times 1$\\
            \hline
            Campus & Campus area & $4.5\times 8$\\
            \hline
            Plaza & Shopping area & $2 \times 1$\\
            \hline
        \end{tabular}
        \label{tab:dataset}
    \end{table}

    \begin{itemize}
        \item \textbf{Pittsburgh} dataset is collected within the city of Pittsburgh with our data-collection platform, which contains a Velodyne-16 LiDAR scanner, Xsens MTI-300 inertial measurement units, and GNSS position systems.
        The collected areas (open street, residential areas, commercial buildings, etc.) contain $50$ trajectories with a total distance of $120$ km and $158$ overlaps.
        \item \textbf{Campus} dataset is recorded with the same data-collection platform within the campus area of Carnegie Mellon University (CMU).
        The collected data covers $10$ main trajectories throughout the campus, where each trajectory is recorded $8$ times under different conditions (illuminations, directions, etc).
        The total length is around $36$km.
        \item \textbf{KITTI} dataset is a well-known dataset for autonomous driving in urban environments.
        We extract out $10$ (around $15$km in total) trajectories from \textit{KITTI} odometry dataset, and mainly used it to evaluate the generalization ability of our place recognition.
    \end{itemize}
    
    Due to sparse LiDAR scanning, occlusion, and changes in perspective, the same place may be represented by different observations.
    To provide consistent local maps for feature extraction, AutoMerge generates a dense map with traditional LiDAR odometry~\cite{LOAM:zhang2014loam}. This approach has been applied in our previous work~\cite{fusionvlad}.
    For each place, sub-maps are constructed by accumulating LiDAR scans into dense observations and keeping a distance ($40$m) to the vehicle's latest position.
    The sites in the CMU campus and \textit{KITTI} datasets have a maximum of two lanes and large lateral displacement with no exceptions.
    The sites in the Pittsburgh dataset street areas have two to four lanes, which indicates a certain lateral displacement during inverse observation.
    The major differences between the \textit{Campus} dataset and the other datasets is the multiple re-visits over the same segments.
    We extract sub-maps every $5$m with a fixed $50$m radius. Fig.~\ref{fig:dense_map} shows example extracted sub-maps.
    These maps can only be extracted when the relative distance between the vehicle's central point and the keyframe's is $100$m away.
    In this manner, the geometric structures for the same areas will be very similar in both under forward and reverse traversal directions. 
    The above datasets enable us to evaluate  place recognition accuracy against rotational and lateral changes, refine data-association robustness against outlier wrong matches, and test map merging performance under large-scale environments.
    In all of the above datasets, we count the retrieval as successful if the detected candidates are $10$m apart from the ground-truth positions.

    \bigskip
    \noindent \textbf{Evaluation Criteria:} To evaluate the loop closure detection accuracy and map merging performance, we use the following three metrics:
    
    1) \textit{Recalls@Top$N$ Retrievals}: AutoMerge uses the best retrievals for map merging, and accurate place retrieval should be invariant to perspective differences.
    We utilize Top-$1$ recall as the main evaluation metric to analyze the place recognition robustness under changing viewpoints.
    
    2) \textit{Precision-Recall Curve}: recall cannot fully represent the general place recognition ability for global merging, as high false positives will make the map optimization fragile even with high recall.
    To this end, we utilize the precision-recall between different segments to investigate the accuracy of retrievals, and the generalization ability for unknown datasets.
    
    3) \textit{Merging Accuracy}: the above metrics mainly focus on fine-grained place recognition accuracy, and cannot fully encapsulate the performance in map-merging tasks.
    Since localization accuracy analysis (i.e., Mean Squared Error) is not realistic for large-scale merging, especially when odometry drift will be a part of the error, AutoMerge use a simplified merging metric.
    We notice that even limited accurate retrievals ($2\sim 4$) on overlaps can provide accurate map merging results.
    For coarse-grained place retrieval accuracy, we care more about the overlaps' binary retrieval rates, i.e., $0/1$ for found/missed.
    
    \bigskip
    \noindent \textbf{Targeting Methods:}
    To analyze the place retrieval accuracy, we compare the fusion-enhanced descriptor extraction of AutoMerge with other state-of-the-art 3D place recognition learning-based methods:
    PointNetVLAD~\cite{PR:pointnetvlad}, PCAN~\cite{PR:pcan}, LPD-Net~\cite{liu2019lpd}, SOE-Net~\cite{PR:soe}, MinkLoc3D~\cite{LPR:minkloc3d} and SphereVLAD~\cite{seqspherevlad}.
    In all the above methods, we use the same sub-map configuration, i.e. $5m$ distance between keyframes, and $50m$ radius and $0.5m^3$ voxelization for each sub-map as shown in Fig.~\ref{fig:dense_map}.
    For map merging evaluation, we only use Top-1 retrieval to detect overlaps among segments, and apply \textit{Merging Accuracy} to provide quantitative analysis and relative quality demonstration to investigate the merging details.
    Please note that point-based methods usually cannot find overlaps in the reverse traversal direction ($180^{\circ}$).
    For a fair comparison, we store the local features for both forward and reverse directions.
    Given the testing and reference queries, we calculate both distances ($cos(f_{ref}^{forward},f_{test}^{forward})$ and $cos(f_{ref}^{forward}, f_{test}^{reverse})$), and use the minimum as the place feature distance.

    \begin{figure*}[t]
        \centering
        \includegraphics[width=0.9\linewidth]{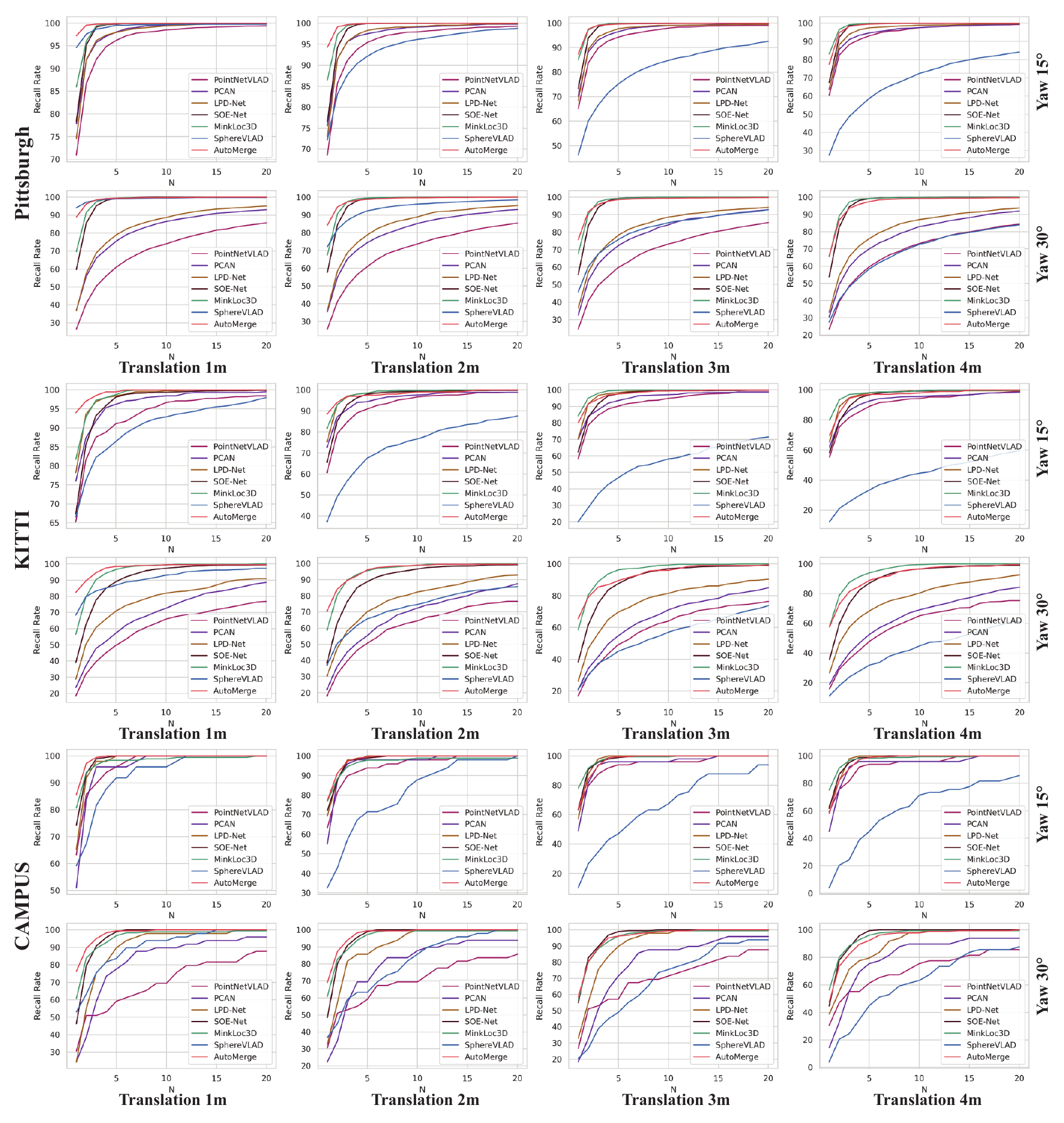}
        \caption{\textbf{Localization results for different viewpoints on different datasets.}
        For each dataset, we pick one segment from the same domain and generate test/reference queries with different yaw angles $[15,30]^{\circ}$ and translational displacement $[1,2,3,4]m$, and then analyze the average recall for top-$20$ retrievals.
        }
        \label{fig:recall_analysis}
    \end{figure*}

To evaluate the generalization of the place recognition and data-association, we used only \textbf{30\%} of the \textit{Pittsburgh} dataset (which is \textbf{20\%} of the total of all three datasets) to train different learning-based methods, and inference over the remaining datasets with the trained models.

    \begin{figure*}[t]
    	\centering
        	\includegraphics[width=0.9\linewidth]{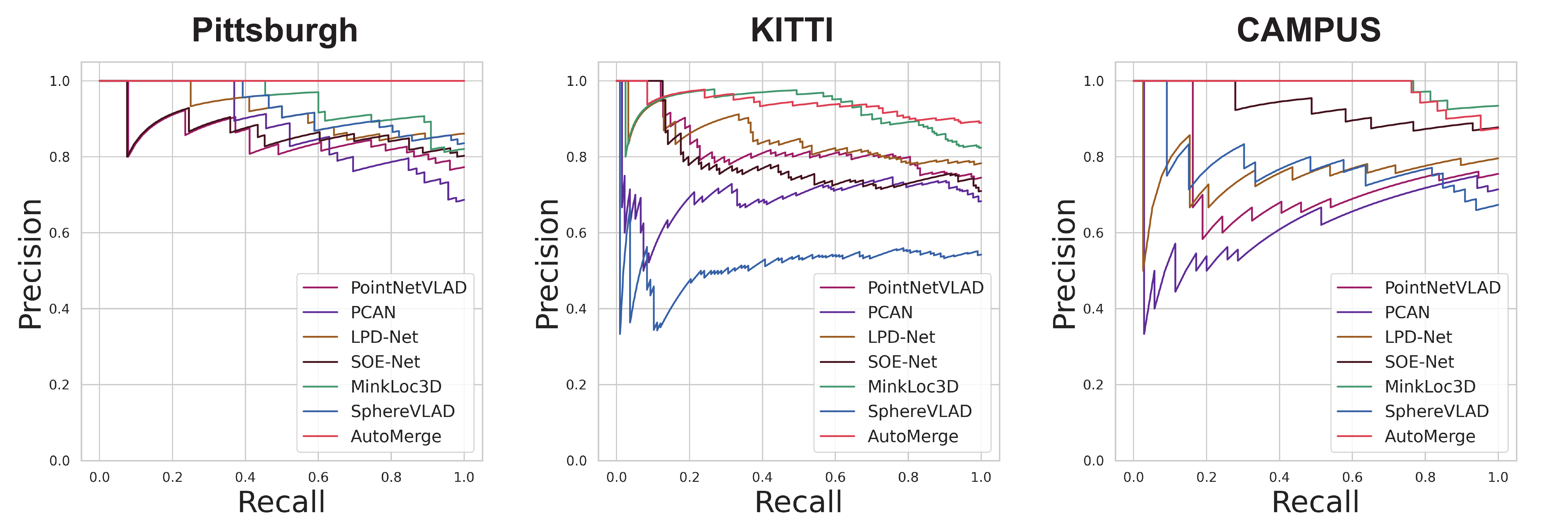}
        	\caption{\textbf{Precision-recall Curves (PR-curve) on three datasets.}
        	 On each dataset, the results are evaluated using testing and reference queries that have a relative translation of $2$m and relative rotation of $15^{\circ}$.
    	}
    	\label{fig:pr_curve}
    \end{figure*}

\section{Experimental Evaluation}
\label{sec:exp_eval}
As is shown in Table.~\ref{tab:dataset} and Fig.~\ref{fig:dataset}, AutoMerge can work with multiple overlaps under city-scale and campus-scale environments, and under various types of scenarios.
Overall, AutoMerge can achieve the best place recognition performance under varying viewpoint differences.
And the map merging results also indicate that data association and incremental merging of AutoMerge are not sensitive to parameter tuning and demonstrate higher generalization potential for new environments.
Compared with other learning-based methods, AutoMerge still shows robust data association ability on all the datasets, even though only trained on \textit{Pittsburgh} dataset.
In this section, we will evaluate the place recognition accuracy, overlap retrieval accuracy, map merging efficiency, and computation efficiency respectively.

\subsection{Place Recognition Results}
\label{sec:PR_results}
\subsubsection{Orientation- and Translation- Tolerance Analysis}
\label{sec:ori_trans}

    We conduct experiments on three datasets to evaluate the robustness of place recognition of different methods. All learning-based methods are trained on tracks $1\sim 15$ of the Pittsburgh dataset.
    As shown in Fig.~\ref{fig:recall_analysis}, we calculate the average Top-$1$ recall between query and reference frames(under translation differences $[1,2,3,4]m$ and yaw orientation differences $[15, 30]^{\circ}$).
    To generate orientation differences, we rotate each query frame by a desired angle and then apply a random noise uniformly sampled from the range $-2.5^{\circ}\sim 2.5^{\circ}$.
    The projection-based method, SphereVLAD\cite{seqspherevlad}, can achieve orientation-invariance, but translation differences will greatly affect the recognition performance.
    Conversely, point-based methods can handle large translation differences but are sensitive to orientation differences.
    We can notice that AutoMerge has the translation-invariant property of point-based methods and the orientation-invariant property of projection-based approaches. This is mainly due to our attention mechanism, which can reweigh the importance of the two branches in the feature extraction model.
    
    On both the \textit{KITTI} and \textit{Pittsburgh} datasets, AutoMerge outperforms both point-based and projection-based methods when subjected to large orientation and translation differences. 
    AutoMerge also shows great generalization ability compared to the single branch point-based and project-based approaches. 
    Moreover, the generalization ability of Automerge indicates that the proposed attention fusion mechanism is not trained to overfit the training dataset.
    We can also notice that MinkLoc3D shows consistent place recognition ability when dealing with significant translation noise.
    However, the same with other point-based methods, the performance declines with the increase of the viewpoint variance.
    In Fig.~\ref{fig:pr_curve}, we analyze the PR-curve of different methods over three datasets.
    We can notice that AutoMerge shows better performance than other descriptors in \textit{Pittsburgh} dataset.
    On the other hand, since all the methods are only trained on \textit{Pittsburgh} dataset, there also exists general performance drop for all the learning-based approaches over the rest two datasets.
    
    To investigate the merging ability, we analyze the merging accuracy over the \textit{Pittsburgh} and \textit{Campus} datasets.
    We extract all the overlaps over the two datasets, and analyze the relative recalls and accuracy of the different methods.
    The results are shown in Table.\ref{table:real_pr}.
    As the distance of each submap is around 5$m$, the ability of the model to deal with variant orientation differences under translation differences around $2m \sim 3m$ is of vital importance.
    Automerge takes advantage of PointNetVLAD and SphereVLAD and achieves higher recall, compared with other methods.
    This capability comes from the adaptive feature association, as stated in Section.~\ref{sec:fusion_network}.

    \begin{table}[t]
        \centering
        \caption{Merging Accuracy Analysis}
        \begin{tabular}{ | c | c | c | c | c |}
        \hline
        Method & \multicolumn{2}{c|}{Pittsburgh} & \multicolumn{2}{c|}{Campus}\\ \hline
        & Precision & Recall & Precision & Recall\\ \hline
        PointNetVLAD        & $82.2\%$ & $31.4\%$   & $92.1\%$  & $87.4\%$ \\ \hline
        PCAN                & $82.6\%$ & $61.2\%$   & $94.6\%$  & $89.2\%$ \\ \hline
        LPD-Net             & $89.2\%$ & $65.3\%$   & $98.6\%$  & $90.3\%$ \\ \hline
        SOE-Net             & $94.0\%$ & $69.3\%$   & $99.4\%$  & $93.6\%$ \\ \hline
        MinkLoc3D           & $\mathbf{96.2}\%$ & $77.6\%$   & $\mathbf{100.0}\%$  & $97.3\%$ \\ \hline
        SphereVLAD          & $95.5\%$ & $72.0\%$   & $\mathbf{100.0\%}$ & $93.5\%$ \\ \hline
        AutoMerge (ours)    & $93.7\%$ & $\mathbf{78.5\%}$   & $\mathbf{100.0\%}$ & $\mathbf{98.2\%}$ \\ \hline
        \end{tabular}
        \label{table:real_pr}
    \end{table}

    \begin{figure*}[t]
    	\centering
        \includegraphics[width=0.95\linewidth]{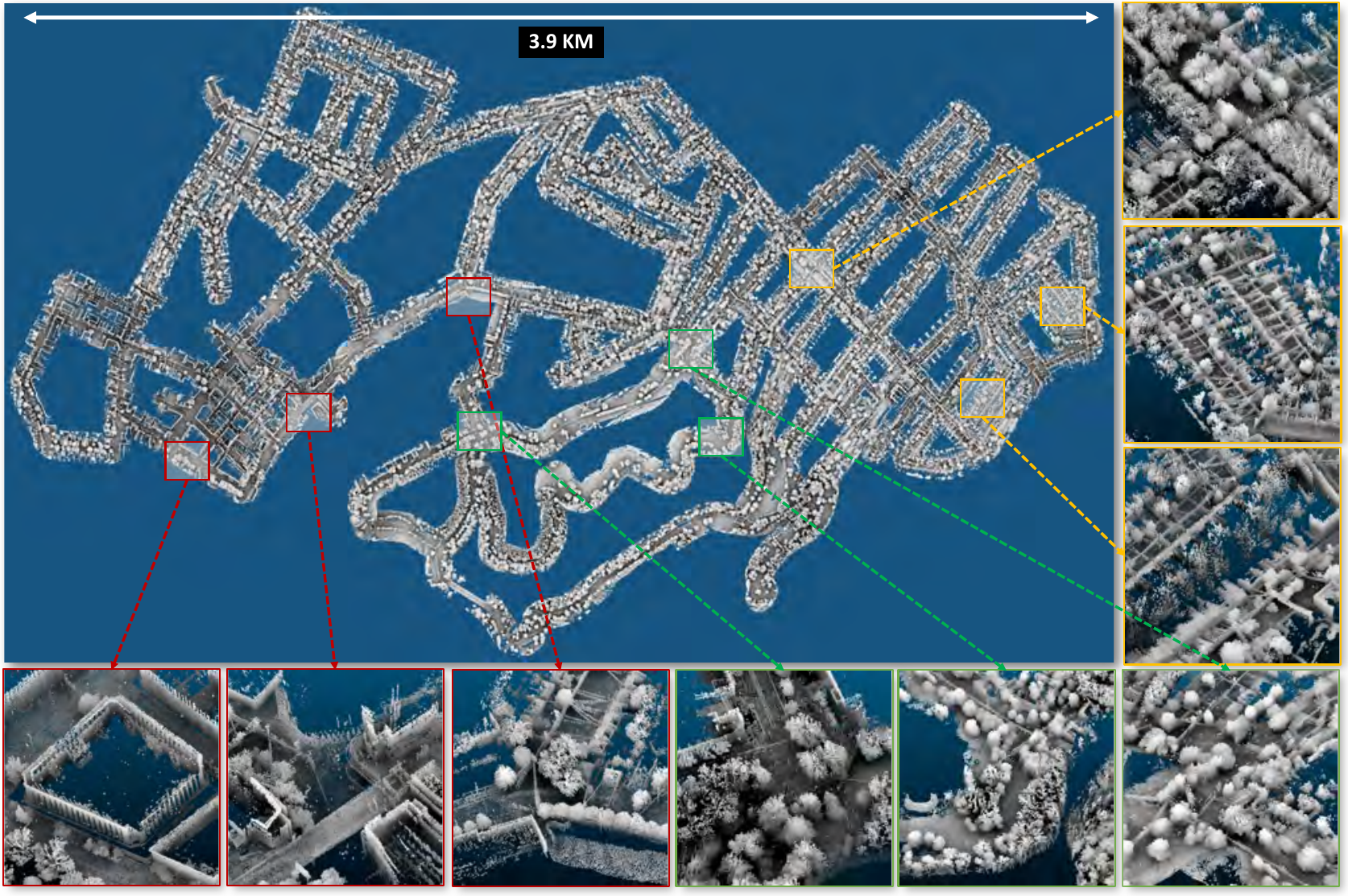}
    	\caption{\textbf{Offline Merging on the \textit{Pittsburgh} Dataset.}
    	The above map is merged using $43$ retrieved segments with limited overlaps from the \textit{Pittsburgh} dataset.
    	We show examples of open-street areas (bounded in red), terrain areas (bounded in green), and residential areas (bounded in yellow).
    	}
    	\label{fig:off_pitts}
    \end{figure*}

    \begin{figure}[t]
    	\centering
        	\includegraphics[width=\linewidth]{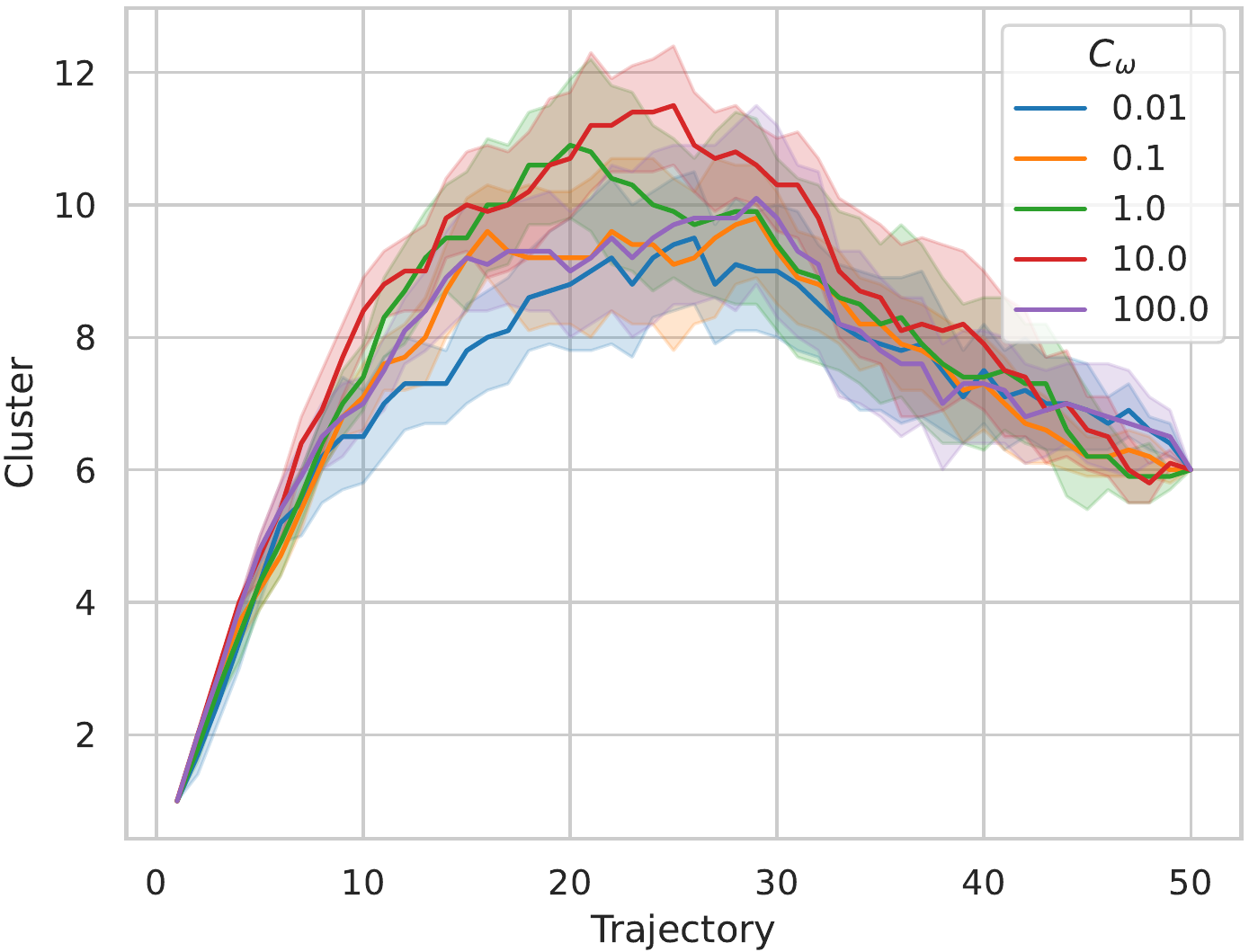}
    	\caption{\textbf{Incremental Clustering under Random Trajectory Order.}
    	We evaluate the incremental clustering performance with different values for hyper-parameter $C_{\omega}$, and we randomly order the trajectory sequences to incrementally update the graphs. 
    	For each $C_{\omega}$, we evaluate the performance $100$ times to analyze the merging trends.
    	}
    	\label{fig:merge_pitt_param}
    \end{figure}

    \begin{figure*}[t]
    	\centering
    	\includegraphics[width=\linewidth]{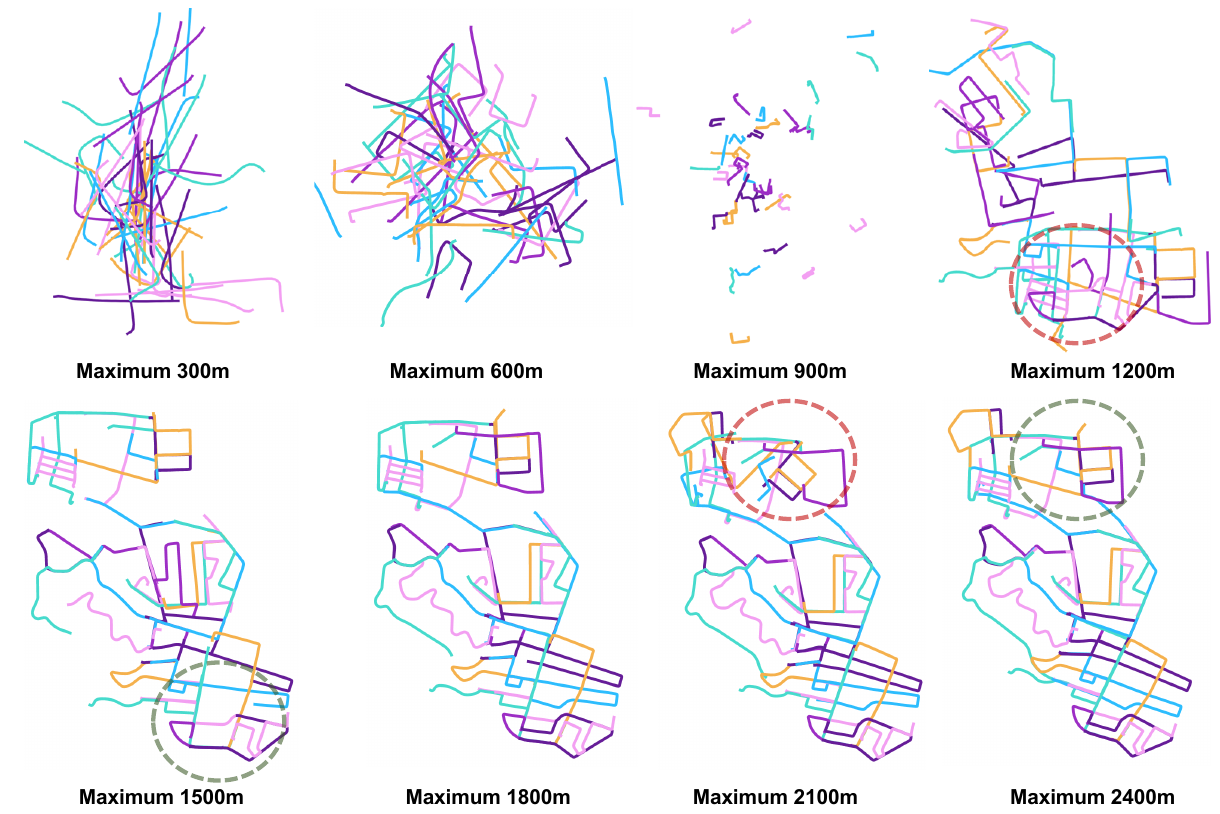}
    	\caption{\textbf{Map Merging with Incremental Expending Trajectories.}
    	This figure shows the incremental merging ability with different maximum segment length limitations, ranging from $300m$ to $2400m$. 
    	Failures due to incorrect matches are shown in red circles, and recovered/updated matches are shown in green circles. AutoMerge shows that it can recover when wrong matches occur during merging.
    	}
    	\label{fig:online_merge_pitt}
    \end{figure*}
    
        \begin{figure}[t]
    	\centering
        	\includegraphics[width=\linewidth]{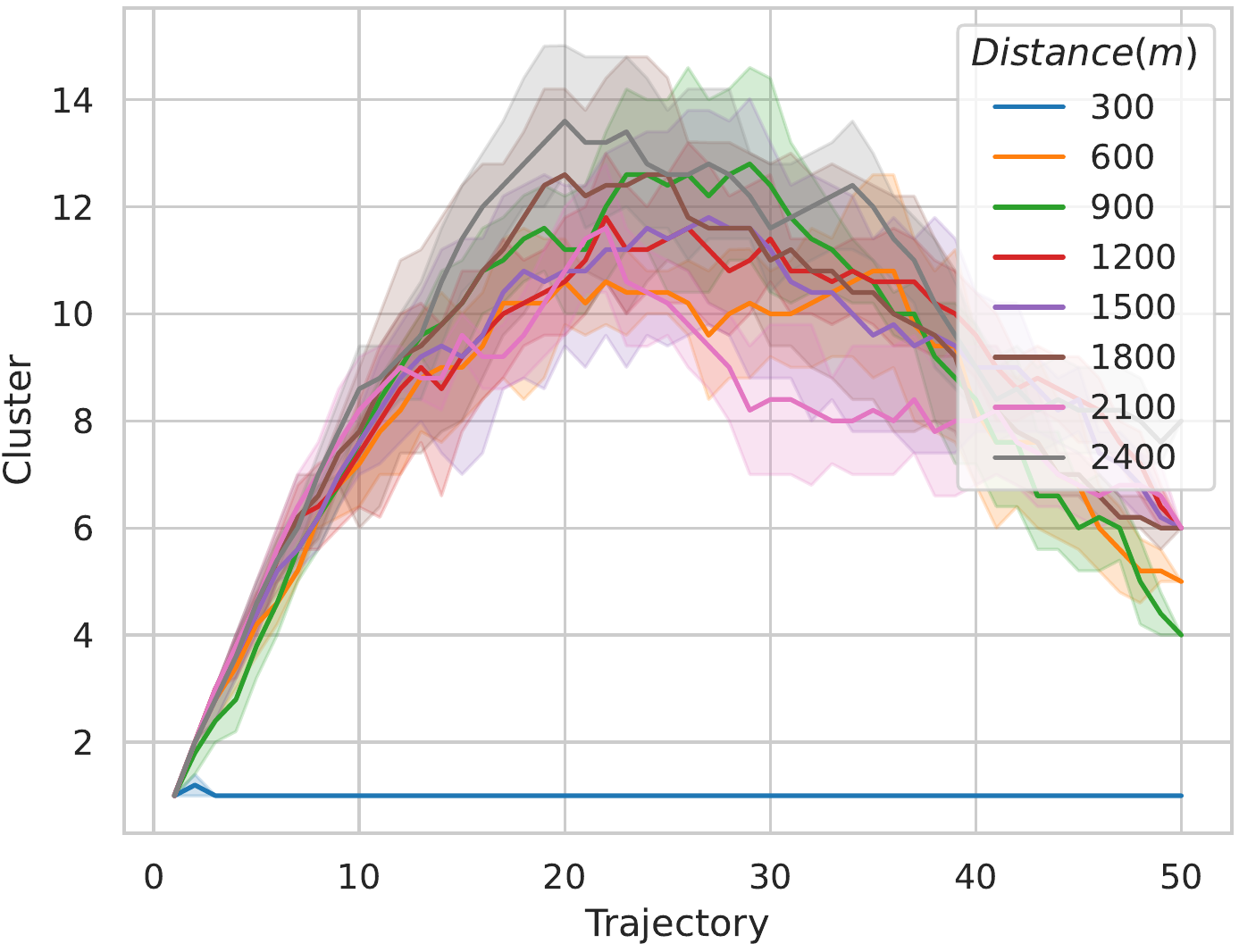}
    	\caption{\textbf{Online Clustering under Changing Distances.}
    	We evaluate the incremental clustering performance under different maximum segment length limitations.
    	We evaluate the cluster results for each distance with $20$ times random order.
    	}
    	\label{fig:merge_pitt_distance}
    \end{figure}

\subsection{Map Merging}
\label{sec:im_cluster}

So far, we have investigated the place recognition results between paired segments.
In this subsection, we consider the multi-segments offline/online map merging task on both the \textit{Pittsburgh} and \textit{Campus} datasets.

\subsubsection{Merging on \textit{Pittsburgh}}
\label{sec:pitt_merge}
In offline merging, we assume all the segments of interest have already been recorded, and AutoMerge can obtain the poses and features over all trajectories at test time.
Based on the relative connections among all the segments, AutoMerge can build the weighted graphs and cluster them into different sub-groups.
In Fig.~\ref{fig:off_pitts}, we evaluate the merging performance over different zones of the \textit{Pittsburgh} dataset.
We can notice that sub-maps in each zone have converged into one consistent large map.
However, not every segment has confident overlaps with other trajectories. 
Those segments with few interactions will be temporarily considered outliers.
For areas with multiple segment overlaps, AutoMerge can also detect the potential connections while ignoring relative viewpoint differences.
This property allows AutoMerge to have robust pose estimation with one-shot visits.
As shown in Table.~\ref{table:real_pr}, the merging performance is robust even in unknown environments.
When tested on the \textit{Pittsburgh} dataset, our model is trained on segments $1\sim10$ giving it $13\%$ dataset coverage. 
This training set only contains areas around Carnegie Mellon University.
The final merging results do not show a significant performance drop over the rest of the datasets, which contains varying terrain, open streets, and residential areas.
Because of its high generalization ability, AutoMerge does not require much data for training.

We also analyzed the robustness of offline clustering in the scenario where AutoMerge is given segments in a randomly selected order.
In Fig.~\ref{fig:merge_pitt_param}, we merged the segments for different values of parameter $C_{\omega}$.
For each parameter, we use a randomly generated $50$ segment streaming order, and calculate the corresponding clustering trends.
The results show that under all cases, AutoMerge can merge \textit{Pittsburgh} segments into $6$ major clusters, and the biggest cluster contains $43$ segments, as shown in Fig.~\ref{fig:off_pitts}.
From this, we can notice that the final clusters are not affected by the segment streaming order and the constant parameter $C_{\omega}$.

For incremental merging on the \textit{Pittsburgh} dataset, we assume all the segments are streamed incrementally.
In this case, at the early merging stage, we can only observe partial trajectories, and wrong matches are unavoidable with these short-term observations.
Using the incremental clustering method depicted in Sec.~\ref{sec:spectral}, AutoMerge can incrementally update the cluster property among segments.
Fig.~\ref{fig:online_merge_pitt} shows the incremental merging results over different maximum segment distances ($300m \sim 2400m$).
Fig.~\ref{fig:merge_pitt_distance} shows the total cluster size under different maximum segment distances.
In the $300m$ and $600m$ cases, the clusters have primarily merged into one cluster.
This is because AutoMerge cannot distinguish different sub-groups when all the connections are weak.
Beginning at the $900m$ distance, partial local overlaps are detected, and all $50$ segments are divided into $4\sim 6$ clusters during the merging procedure.
However, we can notice that not all cases can divided to $6$ clusters as we observed in the offline version of this task.
This is mainly caused by wrong connections between partial observations as indicated in Fig.~\ref{fig:online_merge_pitt}.
We highlight the wrong matches in red circles. These temporary outliers can break the global map as shown in the $1200m$ and $2100m$ cases.
But such failures can be quickly recovered from as shown in the green circles in the $1500m$ and $2400m$ cases. 
In the above experiments we also analyze how the streaming order affects merging for each maximum segment distance. 
The results show that AutoMerge's clustering ability is invariant to the streaming order.

\begin{figure*}[t]
	\centering
	\includegraphics[width=\linewidth]{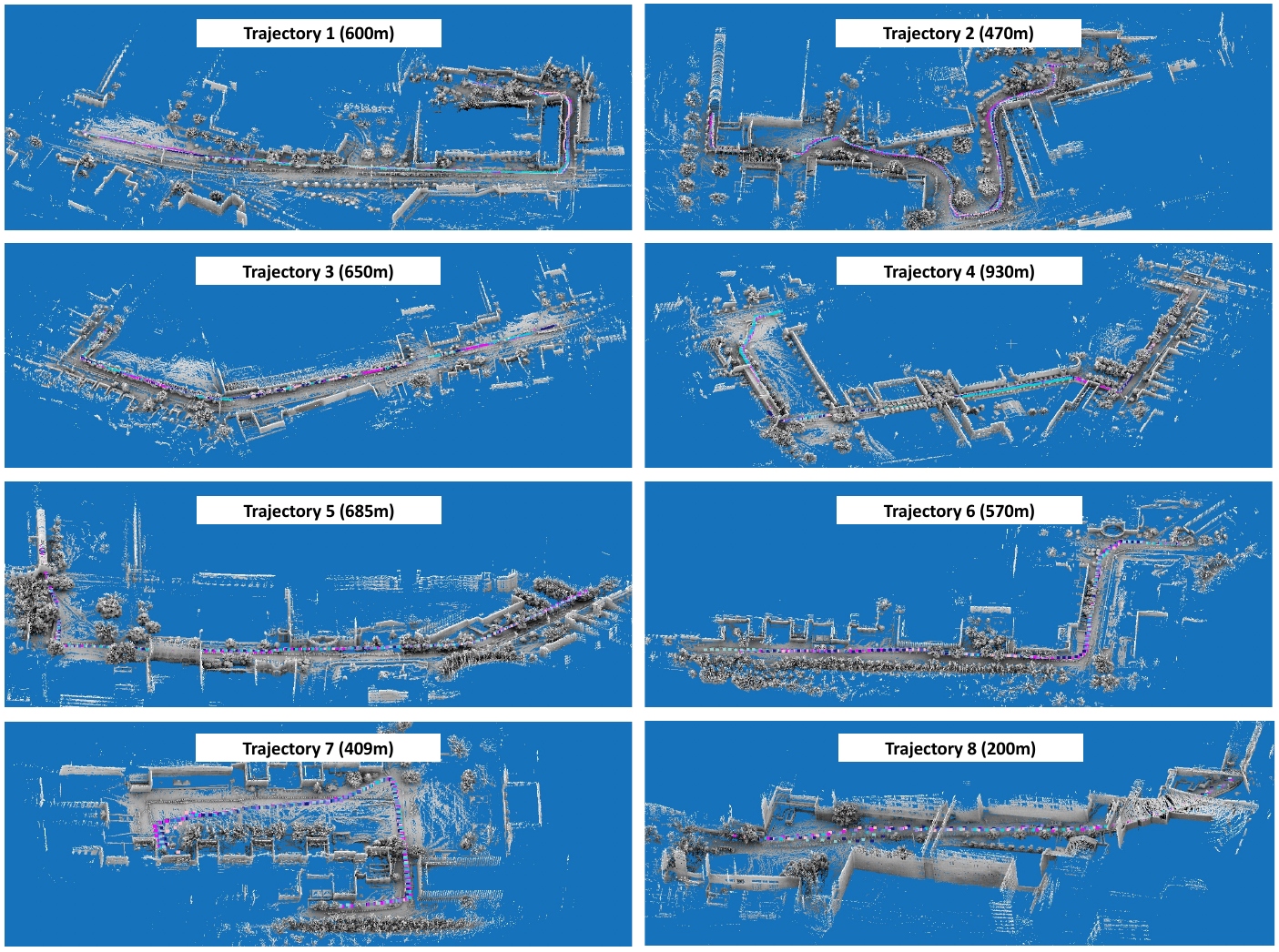}
	\caption{\textbf{Merging over multiple revisits using the \textit{Campus} dataset.}
	The robot revisits $8$ different campus scenarios $8$ times under different traversal directions and illuminations. The above subplots show the map merging results. The colored points indicate the merged segments from all visits.
	}
	\label{fig:offline_campus}
\end{figure*}

\subsubsection{Merging on \textit{Campus}}
\label{sec:cmu_merge}

For the \textit{Campus} dataset, we consider map merging in multi-session scenarios, where each area will be revisited multiple times, with the goal being achieving long-term autonomy.
We chose $8$ scenarios from the campus areas with sufficient temporal differences (from $3\sim 5$ days), and each trajectory is revisited $8$ times with different traversal directions (forward/reverse) and illumination (day/night) conditions.
As shown in Fig.~\ref{fig:offline_campus}, for each segment, we use a one-time visit as the reference map, and the rest of the visits as new queries.
AutoMerge can automatically detect the loop closures between query and test keyframes through our invariant place descriptor and adaptive detection mechanism.
Without any initial estimation, all segments over the same path are able to be transformed into one consistent map.
The final refinements are conducted by Iterative Closet Point (ICP).
However, due to dynamic objects and other sources of noise that occur in multiple visits, noisy merging will occur, especially in confined areas.
This problem is most prevalent in segments $7$ and $8$, which contain lots of dynamic walking pedestrians within confined campus areas, and consequently the merged global map contains lots of merging noise.
Improvements can be made by excluding dynamic objects and using accurate General-ICP~\cite{GICP}, but those methods require additional computation cost.

To better examine the merging performance in multi-session revisits, we visualize the data association for segment $1$ as shown in Fig.~\ref{fig:merge_cmu_trj1}.
Different segments are drawn with different colors, and red links indicate the inner connections between them.
To simplify the visualization, we did not draw all of the links between all of the pairs.
The omnidirectional camera shows the appearances of the same area under different conditions. 
The bottom figures show the difference matrices when comparing the four test segments with one Forward-Day query segment. 
The stable data-association indicates the robustness of AutoMerge in multi-session revisits.

\begin{figure*}[t]
	\centering
    	\includegraphics[width=\linewidth]{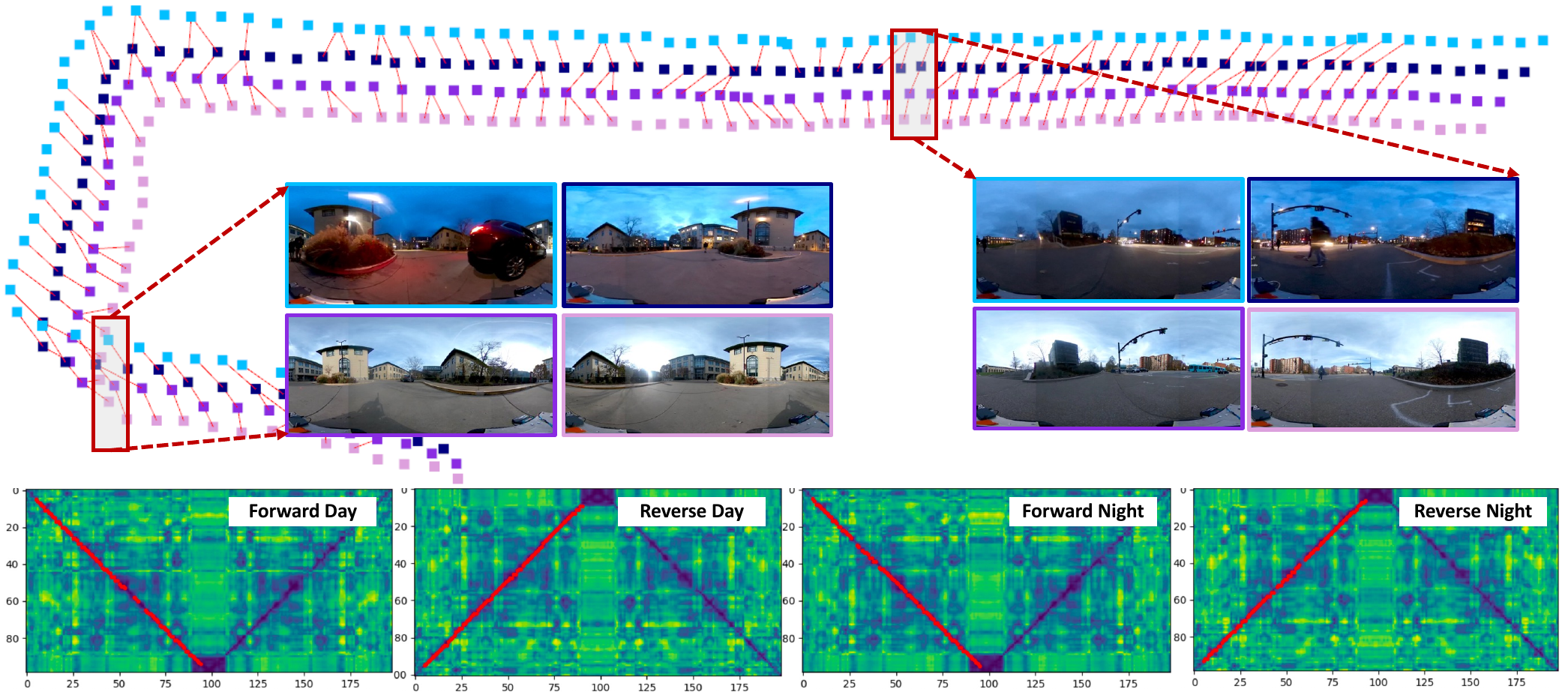}
	\caption{\textbf{Data association on the \textit{Campus} dataset.} 
	Different segments are highlighted in different colors. For the same area we draw the relative data-association among keyframes.
	The omnidirectional camera images show the perspective differences over different revisits.
	The bottom figures show the difference matrices for all cases.
	}
	\label{fig:merge_cmu_trj1}
\end{figure*}

   \begin{figure}[t]
        \centering
        \includegraphics[width=\linewidth]{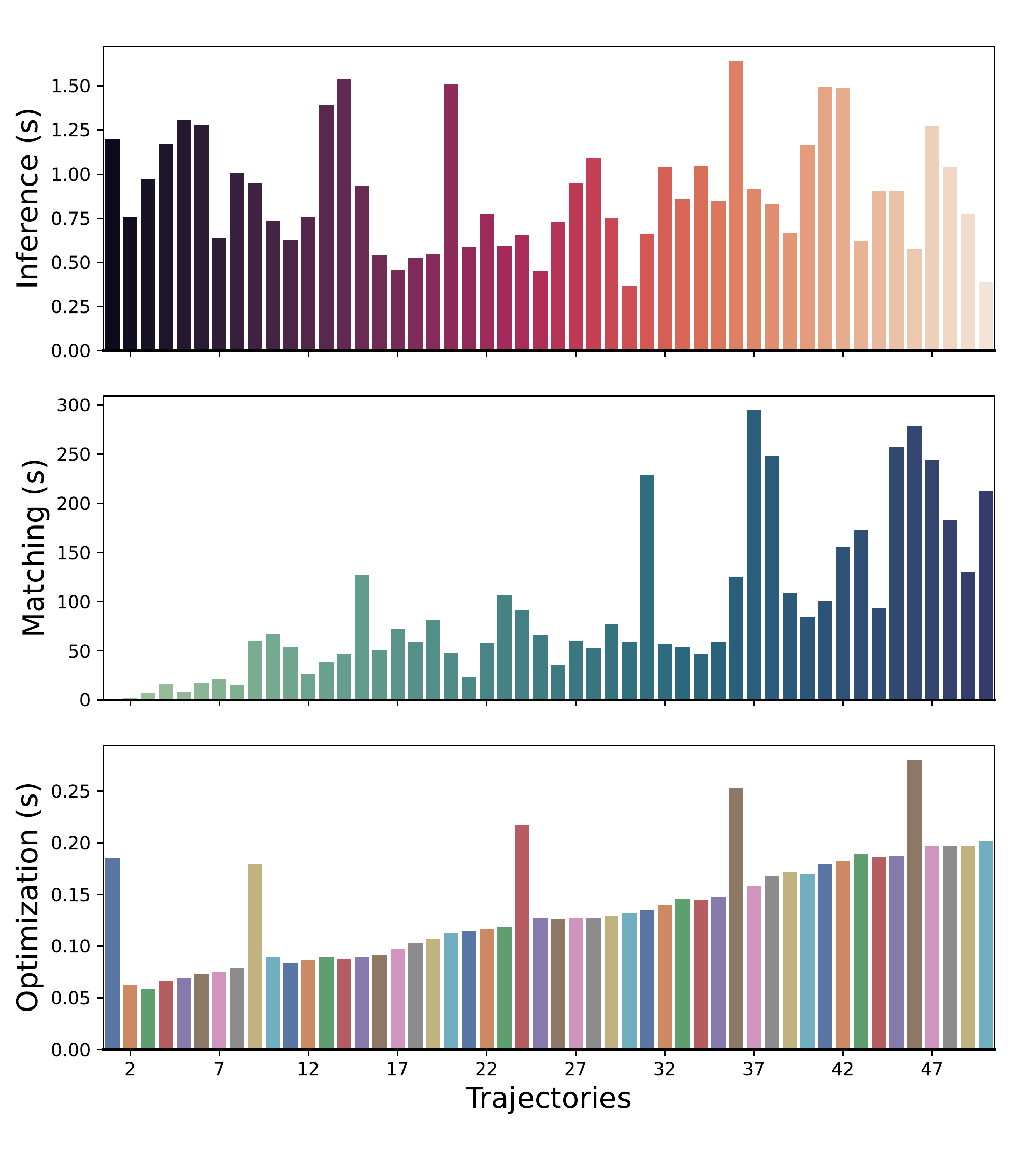}
        \caption{\textbf{Efficiency and Storage Analysis for \textit{Pittsburgh} dataset.}
        (i) The first row shows the inference time for the $50$ segments -- the distances for each segment range from $3km$ to $10km$.
        (ii) The second row shows the highest matching time; every new segment will try to match with all the previous queries.
        (iii) The third row shows the map optimization time.
        }
        \label{fig:time_analsis}
    \end{figure}
    
    \begin{table}[t]
    \centering
    \caption{Comparison of time, GPU memory (Megabyte), and feature size requirements of different methods.
    }
    \begin{tabular}{ |c | c | c | c |}
    \hline
    Method      & GPU (MB)   & Time (ms) & Feature Size  \\ \hline
    PointNetVLAD~\cite{PR:pointnetvlad} & $1,228$ & $4.56$ & $256$ \\ \hline
    PCAN~\cite{PR:pcan} & $7,686$ & $77.06$ & $256$ \\ \hline
    LPD-Net~\cite{liu2019lpd} & $2,578$ & $80.40$ & $256$ \\ \hline
    SOE-Net~\cite{PR:soe} & $3596$ & $94.79$ & $1024$ \\ \hline
    MinkLoc3D~\cite{LPR:minkloc3d} & $1246$ & $15.05$ & $256$\\ \hline
    SphereVLAD~\cite{spherevlad} & $1,069$ & $2.81$ & $512$ \\ \hline
    AutoMerge (ours) & $1,266$ & $13.10$ & $1024$ \\ \hline
    \end{tabular}
    \label{table:time_efficiency}
\end{table}

\subsection{Time and Storage Analysis}
    \label{sec:exp_runtime}
    
    In this section, we compare the proposed method with the current state-of-the-art in learning-based 3D place recognition on both public and self-recorded datasets. 
    To generate our datasets, we designed a data recording mobile platform.
    All the experiments are conducted on an Ubuntu 18.04 system with Nvidia RTX2060 GPU cards and $64$G RAM.
    Table.~\ref{table:time_efficiency} shows the memory usage, inference time, and feature size for all the compared place descriptor methods. 
    Compared with other methods, AutoMerge utilizes less GPU memory and has lower inference time with small storage requirements, which indicates that AutoMerge can be easily combined with current embedded systems.
    
    We further investigate the time efficiency of the incremental merging procedure; in  Fig.~\ref{fig:time_analsis} we analyze the time usage during the incremental map merging for the \textit{Pittsburgh} dataset.
    As we can see, with the AutoMerge framework, both feature extraction and map optimization are efficient, but data association is time-consuming.
    AutoMerge can infer a $5\sim 10 km$ trajectory within $2s$, and optimize the global map within $0.5s$, but data association time ranges from $4s$ to $290s$.
    This is mainly due to the computational complexity of sequence matching~\cite{2012SeqSLAM}, which is $O(n^2)$ where $n$ is the number of keyframes.
    The complexity increases with the reference map scale.
    Since the main procedure in sequence matching is the matrix multiplication operation, one solution for this problem is to apply CUDA-based sequence matching to reduce complexity.

\begin{figure}[t]
	\centering
    	\includegraphics[width=\linewidth]{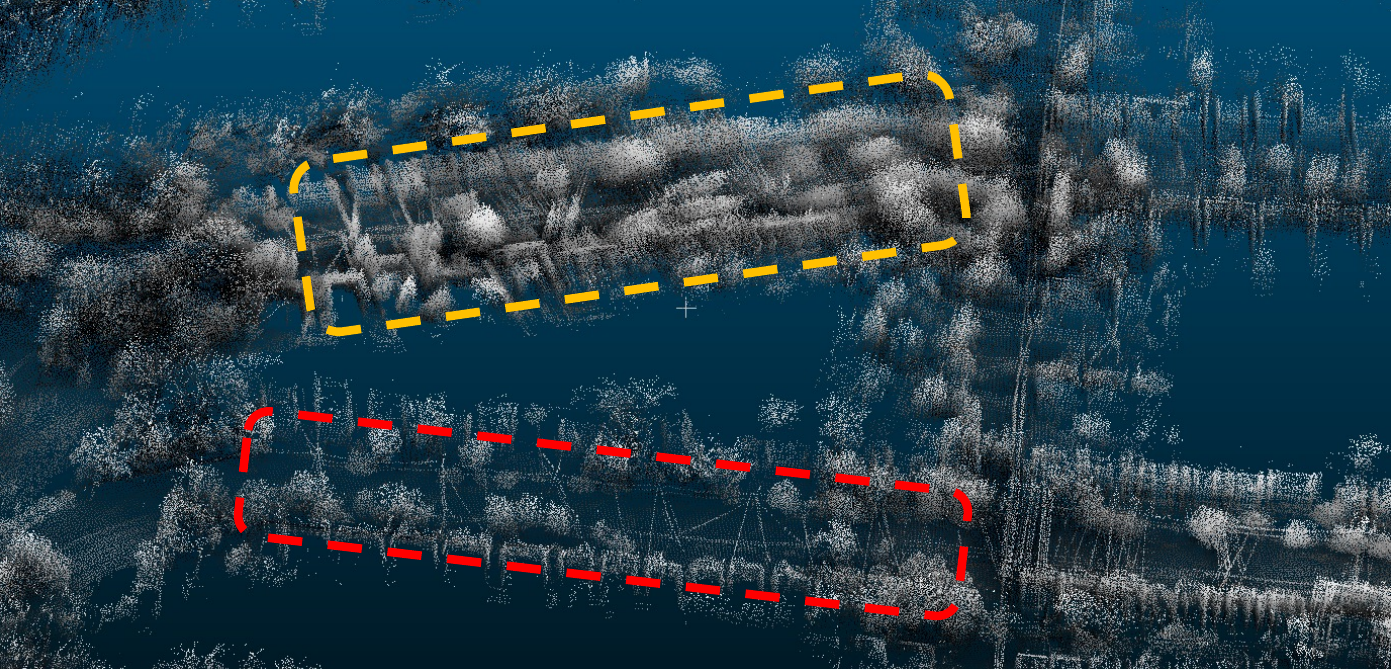}
	\caption{\textbf{Noise and Sparse points in Map Merging.}
	The red circle shows the sparse local map when driving very fast. The yellow circle shows the noisy local map when encountering dynamic objects. These cases will introduce noise to the local place descriptor, which will affect the merging results.
	}
	\label{fig:merge_noise}
\end{figure}

\section{Discussion \& Limitations}
\label{sec:discussion}
    As shown in the above analysis, AutoMerge can provide robust map merging for city-scale and campus-scale environments without any initial estimation.
    This framework can provide offline/online merging for single- and multi-agent systems while ignoring viewpoint and temporal differences common in real-world mapping scenarios.
    However, AutoMerge also has the following limitations.
    
    AutoMerge heavily utilizes generated dense local maps. Thus, the place recognition accuracy is determined by the stability of these local maps.
    As shown in Fig.~\ref{fig:merge_noise}, when the agent is moving too fast (red circle) or there exist too many dynamic objects (yellow circle), the noise and sparse local maps negatively impact the merging procedure.
    Such observations will introduce uncertainty in the extracted place descriptor and indirectly affect the final merging performance.
    
    Besides the noise and sparsity, the extracted place descriptor is also sensitive to confined environments.
    In indoor areas, tunnels, and underground environments, the generated LiDAR map is constrained within a relatively small space compared to outdoor environments.
    In these cases, distinguishable features cannot be easily extracted from either point-based or spherical projection-based data formats.
    To obtain rich geometries, point meshlization could be a potential solution.

    Additionally, the adaptive loop closure detection relies highly on the sequence matching results, and subsequently, its sequence searching process is the most time-consuming part of AutoMerge. Since the main procedure in sequence matching is the brute-force searching operation, a CUDA-enhanced sequence matching mechanism can further improve searching efficiency, as mentioned in~\cite{cuda_seqslam}.
    
    AutoMerge cannot handle trajectories with limited overlap.
    Since high merging accuracy is our primary goal, only high confidence overlaps are selected as loop closure candidates.
    The major drawback of this mechanism is missed loop closures in trajectories with minimum overlap. These cases usually occur at crossroads where neighbor trajectories only have $1\sim 2$ matched keyframes.
    However, from the standpoint of large-scale merging performance, this principle is necessary since we need to detect potential overlaps within hundreds of kilometers of trajectories; in such a scenario, several wrong short-range matches will crash the entire system.

    Furthermore, AutoMerge cannot handle degradation areas without GPS assistance, such as highways, long tunnels, etc.
    We have tested AutoMerge in another campus-scale dataset collected within a shopping plaza in the City of Shenzhen.
    The collected data includes $8$ trajectories covering the commercial streets and underground passages and we select three trajectories that share overlapped areas in an underground passage.
    Since the indoor environment is relatively narrow in space, we maintained the same model parameters with other datasets while adjusting the radius from $50$m to $20$m in submap generation\ref{sec:data_crit}.
    Automerge can successfully detect the correct overlaps but a part of the correspondences within the overlapped areas can be recognized owing to the structural similarity in underground environments.
    As shown in Fig.~\ref{fig:discuss_under_trajs}, transformation matrices generated from the rough alignment exist rotational errors due to the lack of sufficient detected correspondences.
    However, with the help of back-end optimization, the errors introduced by rough alignment are alleviated and the final merged map is represented in Fig.\ref{fig:discuss_under_pcd}.
    The challenges in the degradation areas come from two-folds:
    1) the non-distinguishable place descriptor will reduce the overlap detection accuracy in such areas.
    2) the degradation areas will also be challenging for the odometry estimation, which indirectly affects the key-frame extraction (given that the distance between key-frames is based on the odometry estimation).
    A potential solution is to combine texture-rich visual features into the place descriptor engine as stated in our previous work~\cite{adafusion} and add fuse visual/wheel odometry into the LiDAR SLAM system to reduce the odometry drift in the degradation area.
    Enable AutoMerge under the degradation case is also an inspiring trend; we would like to leave it to our further work.

    In general, the map merging ability of AutoMerge can be further extended with other types of sensors (e.g., new types of LiDAR or visual sensors) and place descriptor extraction methods.
    Because AutoMerge provides a map merging framework, any existing modules within AutoMerge can be replaced to fit the specific properties of other sensors and network structures.

    Finally, data compression can also be a potential research extension for AutoMerge.
    In our experiment, we notice that AutoMerge can use low-resolution ($0.5m$ in voxel) point clouds for roughly large-scale map merging.
    Given the current research progress on point cloud compression~\cite{wiesmann2021ral}, we notice that will be a potential chance for large-scale map sharing under low-bandwidth communication, especially for service robotics, last-mile delivery and autonomous driving.
    Another potential direction is to combine AutoMerge with the increasing requirements of the low-cost visual localization system~\cite{i3dloc,panek2022meshloc}, where AutoMerge can provide the reference meshes/semantics for accurate visual navigation.
\begin{figure*}[t]
    \centering
    \includegraphics[width=\linewidth]{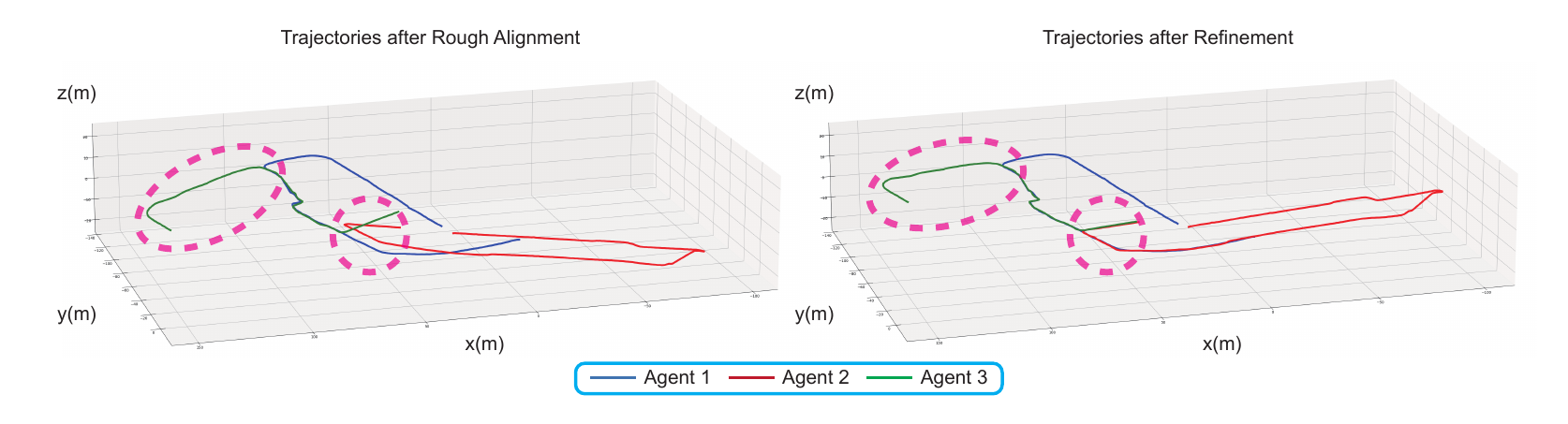}
    \caption{\textbf{Merged trajectories after rough alignment and pose graph refinement.}
    In the underground environment, the transformation matrices calculated from rough alignment exist with obvious rotation errors as shown in the left plot.
    However, these errors can be eliminated by our back-end optimization.
    }
    \label{fig:discuss_under_trajs}
\end{figure*}

\begin{figure}[ht]
    \centering
    \includegraphics[width=\linewidth]{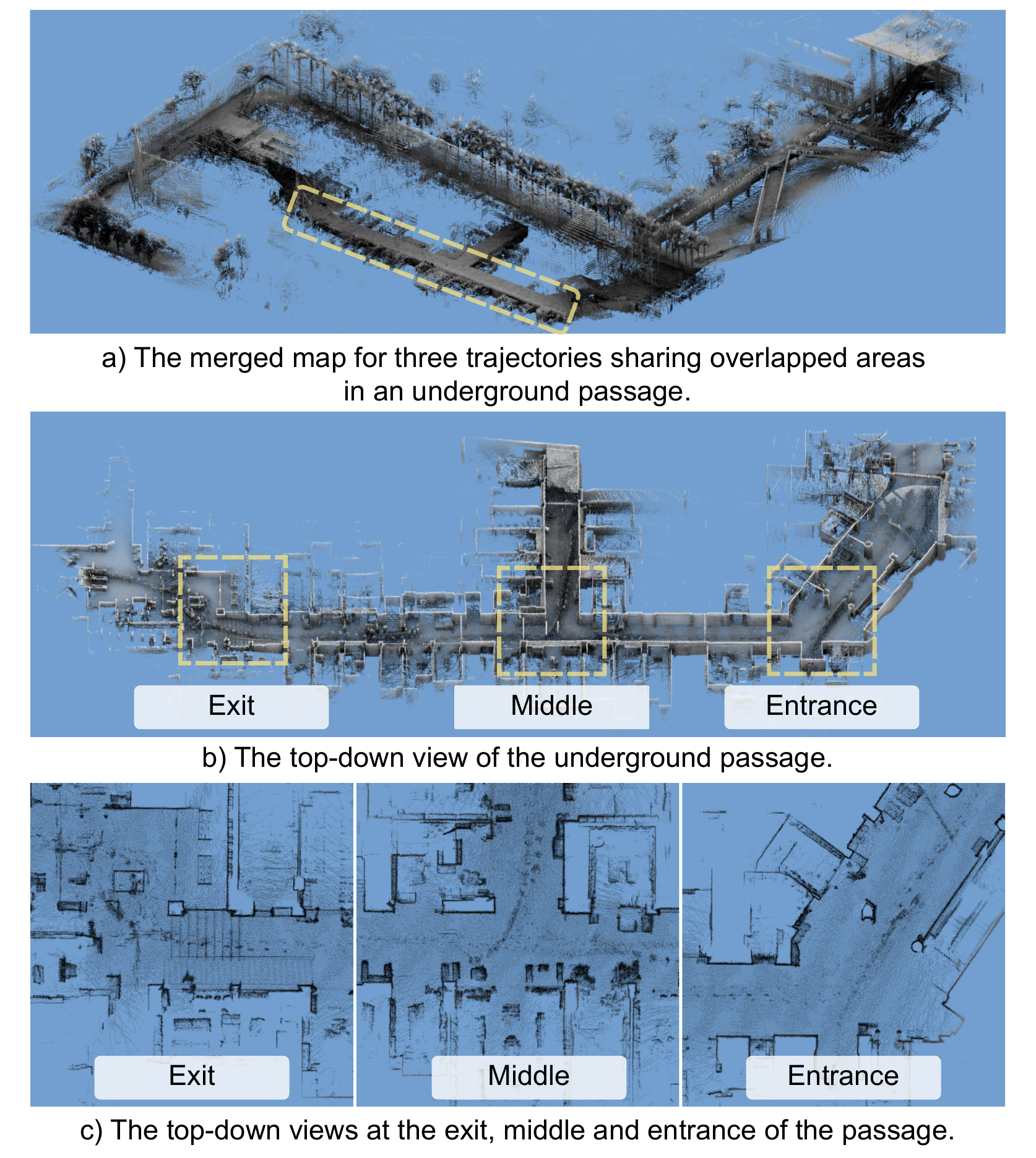}
    \caption{\textbf{Merged point cloud in the underground environment.}
    a) and b) visualize the merged map of three trajectories and the overlapped area(i.e. underground passage) respectively;
    c) shows the point cloud at the overlapped areas of two trajectories located at the exit and entrance of the passage and the overlapped area of three trajectories located in the middle.
    }
    \label{fig:discuss_under_pcd}
    \end{figure}
\section{Conclusions}
\label{sec:conclusions}
    In this paper, we proposed AutoMerge, the first real-world automatic merging system for large-scale 3D mapping.
    AutoMerge can automatically detect the relative overlaps between segments due to its viewpoint-invariant place recognition ability and enhance the matching results with sequence matching.
    Despite the complicated city-scale environments and similar-looking 3D areas under different scenarios, AutoMerge provides highly accurate data associated with our adaptive loop closure detection module.
    Finally, AutoMerge can successfully merge sub-segments given in non-sequential order using the incremental merging module.
    The above properties make AutoMerge suitable to merge large-scale maps, such as city-scale, campus-scale, and subterranean environments.
    
    The results on both public and self-recorded datasets show that our place retrieval ability notably outperforms all state-of-the-art methods in 3D loop closure detection.
    Because of its high recall rates and incremental merging ability, AutoMerge seems like a promising method to use on various real-world datasets.
    Our method can work with limited computational resources and storage space, making it extremely suitable for low-cost robots in large-scale map merging tasks.
    In future works, we will target the current limitations of our method and make this code publicly available.

\bibliographystyle{IEEEtran}
\bibliography{bible}
\vspace{-1cm}
\begin{IEEEbiography}[{\includegraphics[width=1in,height=1.25in,clip,keepaspectratio]{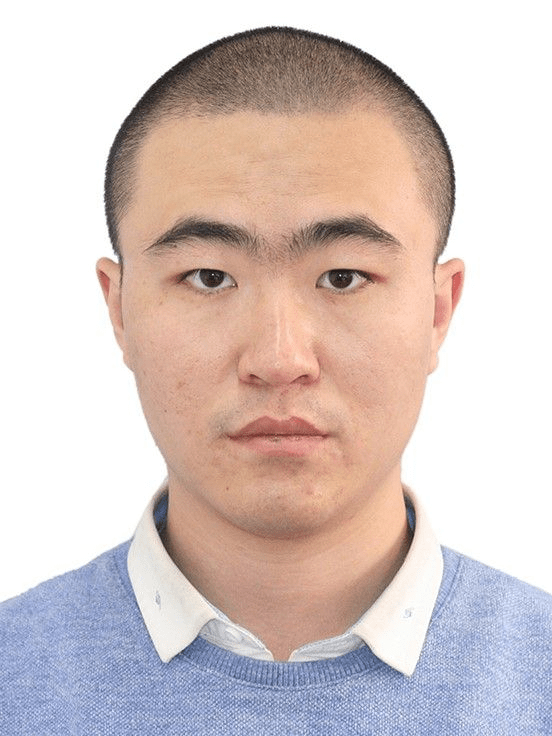}}]
    {Peng Yin} received his Bachelor's degree from Harbin Institute of Technology, Harbin, China, in 2013, and his Ph.D. degree from the University of Chinese Academy of Sciences, Beijing, in 2018.
    He is currently an Assistant Professor at City University of Hong Kong, China.
    His research interests include LiDAR SLAM, Place Recognition, 3D Perception, and Reinforcement Learning. Dr. Yin has served as a Reviewer for several IEEE Conferences ICRA, IROS, ACC, RSS.
\end{IEEEbiography}

\begin{IEEEbiography}[{\includegraphics[width=1in,height=1.25in,clip,keepaspectratio]{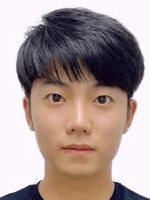}}]
    {Shiqi Zhao} received his Bachelor's degree from Dalian University of Technology, Dalian, China, in 2018, and his Master's degree from the University of California San Diego, U.S., in 2020.
    He is currently working as a research assistant at City University of Hong Kong.
    His research interests include Place Recognition, 3D Perception, and Deep Learning.
\end{IEEEbiography}

\begin{IEEEbiography}[{\includegraphics[width=1in,height=1.25in,clip,keepaspectratio]{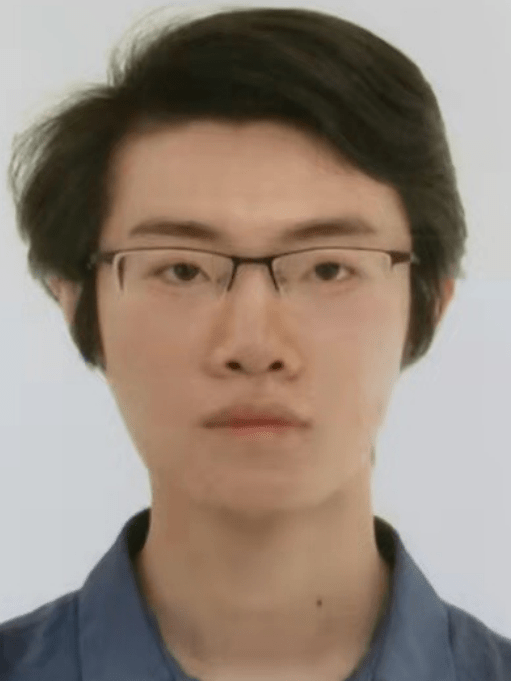}}]
    {Haowen Lai} received his B.E. degree in control science and engineering from Tongji University, Shanghai, China, in 2019. He is currently pursuing an M.S. degree from Tsinghua University in Beijing, China. His research mainly covers 3D localization and perception, SLAM, Place Recognition, and their application in robotics. 
    He is trying to apply computer vision and machine learning techniques to robotics so as to make robots more intelligent in understanding the environment.
\end{IEEEbiography}

\begin{IEEEbiography}[{\includegraphics[width=1in,height=1.25in,clip,keepaspectratio]{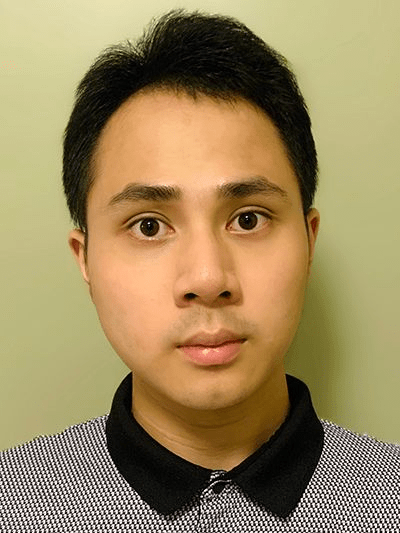}}]
    {Ruohai Ge} received his B.S. degree in Electrical Engineering with a double major in Robotics from Carnegie Mellon University in 2016. He is currently working on a Master's degree in Robotics within the Robotics Institute at Carnegie Mellon University. His research interests include visual localization and robotic related software infrastructure. 
\end{IEEEbiography}



\begin{IEEEbiography}[{\includegraphics[width=1in,height=1.25in,clip,keepaspectratio]{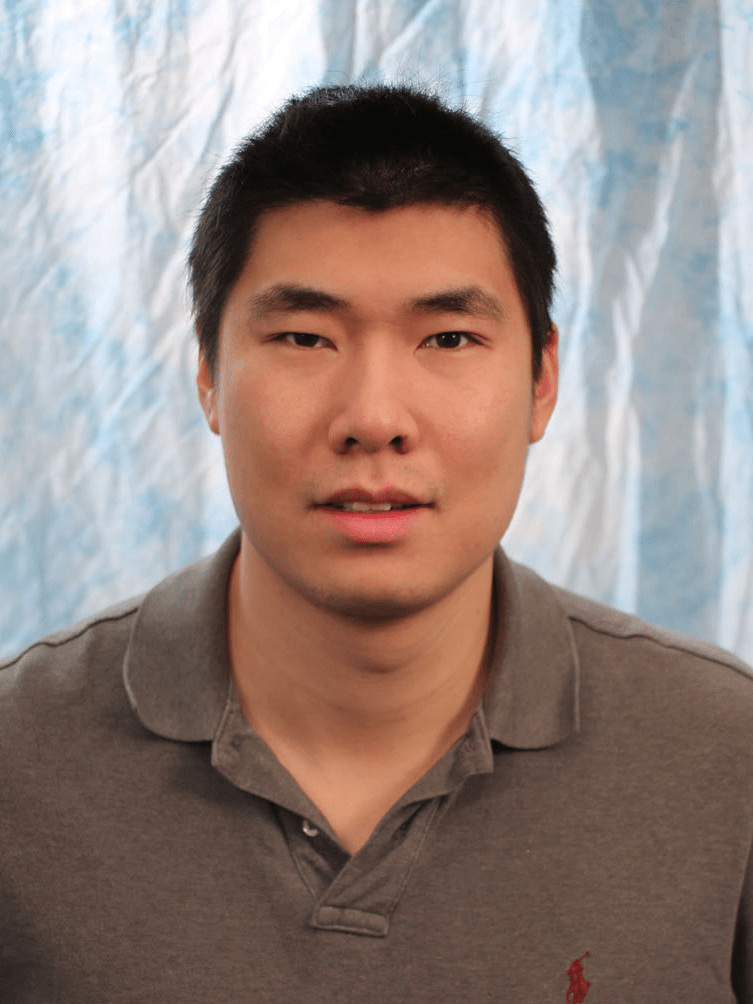}}]
    {Ji Zhang} received his Ph.D. in Robotics from Carnegie Mellon University in 2017.
    Ji Zhang is a Systems Scientist at the Robotics Institute at Carnegie Mellon University, where he leads in the development of a series of autonomous navigation algorithms. His work was ranked \#1 on the odometry leaderboard of KITTI Vision Benchmark between 2014 and 2021. 
    He founded Kaarta, Inc, a CMU spin-off commercializing 3D mapping \& modeling technologies, and stayed with the company for 4 years as chief scientist.
    His research interests are in robotic navigation, spanning localization, mapping, planning, and exploration.
\end{IEEEbiography}

\begin{IEEEbiography}[{\includegraphics[width=1in,height=1.25in,clip,keepaspectratio]{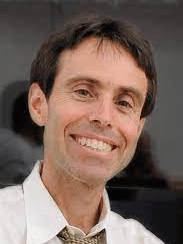}}]
    {Howie Choset} received his B.S. Eng. degree in computer science and his B.S. Econ. degree in entrepreneurial management from the University of Pennsylvania (Wharton), Philadelphia, PA, USA, in 1990. He received M.S. and Ph.D. degrees in mechanical engineering from California Institute of Technology (Caltech), Pasadena, CA, USA, in 1991 and 1996, respectively.
    He is currently a Professor of Robotics at Carnegie Mellon University, Pittsburgh, PA, USA. His research group reduces complicated high dimensional problems found in robotics to low-dimensional simpler ones for design, analysis, and planning. 
\end{IEEEbiography}

\vspace{-1cm}
\begin{IEEEbiography}[{\includegraphics[width=1in,height=1.25in,clip,keepaspectratio]{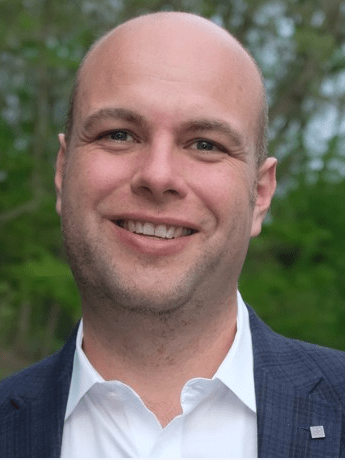}}]
    {Sebastian Scherer} received his B.S. in Computer Science, M.S. and Ph.D. in Robotics from CMU in 2004, 2007, and 2010. 
    Sebastian Scherer is an Associate Research Professor at the Robotics Institute at Carnegie Mellon University. His research focuses on enabling autonomy for unmanned rotorcraft to operate at low altitude in cluttered environments. He is a Siebel scholar and a recipient of multiple paper awards and nominations, including AIAA@Infotech 2010 and FSR 2013. His research has been covered by the national and internal press including IEEE Spectrum, the New Scientist, Wired, der Spiegel, and the WSJ. 
\end{IEEEbiography}

\endgroup
\end{document}